\documentclass[sigconf]{acmart}

\usepackage[utf8]{inputenc} 
\usepackage[T1]{fontenc}    
\usepackage{hyperref}       
\usepackage{url}            
\usepackage{booktabs}       
\usepackage{amsfonts}       
\usepackage{nicefrac}       
\usepackage{microtype}      

\usepackage{amsmath}
\usepackage{subcaption}
\usepackage{graphicx}
\usepackage{tikz}

\usepackage{etoolbox}
\usepackage[ruled]{algorithm2e}
\usepackage{multirow}
\usepackage{hhline}
\usepackage{wrapfig}
\usepackage{comment}

\usepackage{xcolor}

\usepackage{rotating}

\usepackage{xspace}
\usepackage{units}

\definecolor{myred}{rgb}{0.8,0,0}
\definecolor{mygreen}{rgb}{0,0.6,0}
\definecolor{myblue}{rgb}{0,0,0.7}

\DeclareMathOperator{\avg}{mean}

\AtBeginDocument{%
  \providecommand\BibTeX{{%
    \normalfont B\kern-0.5em{\scshape i\kern-0.25em b}\kern-0.8em\TeX}}}

\setcopyright{acmcopyright}
\copyrightyear{2023}
\acmYear{2023}
\acmDOI{10.1145/1122445.1122456}

\acmConference[]{Feb}{10}{2023}

\begin{document}


\title{The Quality-Diversity Transformer: Generating Behavior-Conditioned Trajectories with Decision Transformers}

\author{Valentin Mac\'e}
\affiliation{%
  \institution{InstaDeep}
  \city{Paris}
  \country{France}
}
\email{v.mace@instadeep.com}

\author{Raphaël Boige}
\affiliation{%
  \institution{InstaDeep}
  \city{Paris}
  \country{France}
}
\email{r.boige@instadeep.com}

\author{Felix Chalumeau}
\affiliation{%
  \institution{InstaDeep}
  \city{Paris}
  \country{France}
}
\email{f.chalumeau@instadeep.com}

 \author{Thomas Pierrot}
 \affiliation{%
   \institution{InstaDeep}
   \city{Boston}
   \country{USA}
 }
 \email{t.pierrot@instadeep.com}

 \author{Guillaume Richard}
 \affiliation{%
  \institution{InstaDeep}
  \city{Paris}
  \country{France}
 }
\email{g.richard@instadeep.com}

\author{Nicolas Perrin-Gilbert}
\affiliation{%
  \institution{CNRS, Sorbonne Universit\'e}
  \city{Paris}
  \country{France}
}
\email{perrin@isir.upmc.fr}

\renewcommand{\shortauthors}{Mac\'e et al.}

\begin{abstract}

In the context of neuroevolution, Quality-Diversity algorithms have proven effective in generating repertoires of diverse and efficient policies by relying on the definition of a behavior space. A natural goal induced by the creation of such a repertoire is trying to achieve behaviors on demand, which can be done by running the corresponding policy from the repertoire. However, in uncertain environments, two problems arise. First, policies can lack robustness and repeatability, meaning that multiple episodes under slightly different conditions often result in very different behaviors. Second, due to the discrete nature of the repertoire, solutions vary discontinuously. 
Here we present a new approach to achieve behavior-conditioned trajectory generation based on two mechanisms: First, MAP-Elites Low-Spread (ME-LS), which constrains the selection of solutions to those that are the most consistent in the behavior space. Second, the Quality-Diversity Transformer (QDT), a Transformer-based model conditioned on continuous behavior descriptors, which trains on a dataset generated by policies from a ME-LS repertoire and learns to autoregressively generate sequences of actions that achieve target behaviors. Results show that ME-LS produces consistent and robust policies, and that its combination with the QDT yields a single policy capable of achieving diverse behaviors on demand with high accuracy.

\end{abstract}





\begin{teaserfigure}
\centering
    \begin{subfigure}{\linewidth}
        \includegraphics[width=\textwidth]{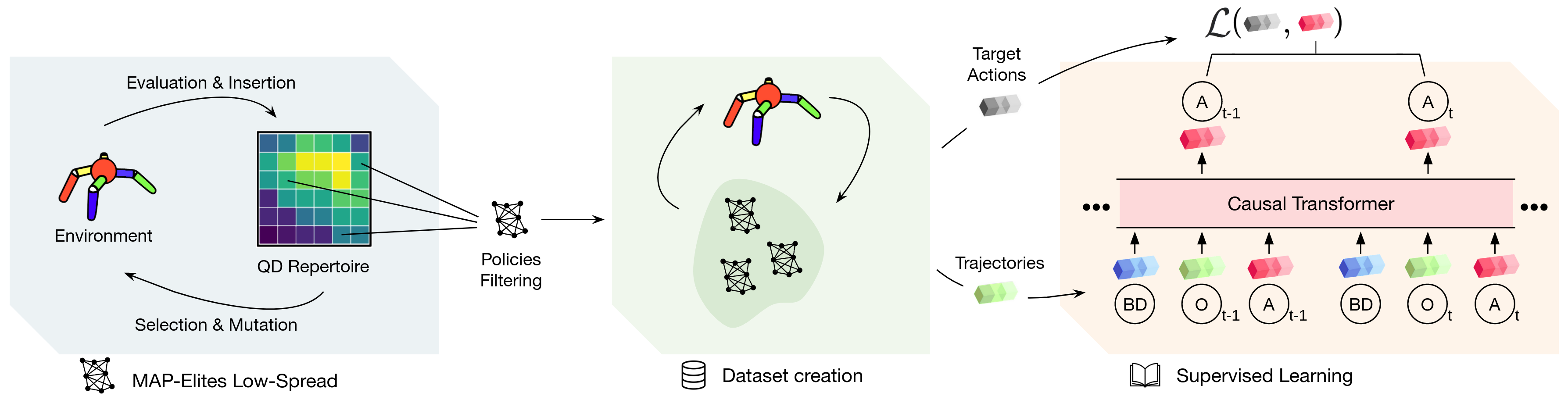}
        \Description{AntTrap final solutions}
    \end{subfigure}
    \caption{High-level concepts of the Quality-Diversity Transformer pipeline. Our full algorithm is composed of three distinct and successive parts: 1. Execution of MAP-Elites Low-Spread, our variation of MAP-Elites that produces stable and consistent policies, 2. Creation of an offline dataset of trajectories by playing episodes with some of the policies produced by MAP-Elites Low-Spread, 3. Supervised training of a causal Transformer that can
    produce a behavior on demand by conditioning on it.
    }
    \label{fig:teaser}
\end{teaserfigure}

    \maketitle

\section{Introduction}
Inspired by nature and its ability to evolve a wide range of diverse and well-adapted organisms, Quality-Diversity (QD) algorithms are a class of evolutionary algorithms that generate diverse and efficient solutions to optimization problems~\citep{pugh2016quality, cully2017quality}.
These methods have been applied in a variety of fields, including robotics~\citep{cully2015robots, fontaine2020quality}, investment~\cite{zhang2020autoalpha}, gaming~\citep{fontaine2019, charity2020baba} and multi-objective optimization~\citep{pierrot2022multi}. They have shown to be particularly useful in problems where the reward signal is sparse or deceptive~\citep{colas2020scaling, pierrot2022diversity}.
In traditional optimization methods, the focus is on finding the single best solution to a given problem. However, in many real-world scenarios, it is important to explore the entire solution space and find diverse efficient solutions that can be used as alternatives in case a single solution fails~\cite{cully2015robots}. This principle is at the core of QD methods and is sometimes referred to as ``illumination'' in opposition to optimization, as it reveals --or illuminates-- a search space of interest~\citep{mouret2015illuminating}, often called the behavior descriptor space (BD space), or simply the behavior space.
Previous work has shown that QD methods are suitable for neuroevolution~\cite{rakicevic2021policy, flageat2022benchmarking} in complex, uncertain domains such as robotic manipulation and locomotion~\cite{colas2020scaling, nilsson2021policy, morrison2020egad}. However, they demonstrate important difficulties in producing policies that are consistent in the behavior space when the initial conditions of an episode change slightly~\cite{flageat2022empirical}.

Moving to another topic related to this work, recent advances in machine learning have led to the emergence of the Transformer~\cite{vaswani2017attention} as a powerful and prevalent model architecture to address various problems, including text generation~\cite{brown2020language}, image processing~\cite{dosovitskiy2020image, carion2020end} and sequential decision making~\cite{reed2022generalist, lee2022multigame}. In particular, the Decision Transformer (DT)~\cite{chen2021decision} performs conditioned sequential decision making in simulated robotics environments by leveraging supervised learning on datasets of trajectories generated by reinforcement learning (RL) policies. Unlike other approaches, its specificity is to condition its decision making process on a desired return 
to be obtained at the end of an episode. After an offline training phase, the DT is able to condition on a target return and to play episodes that achieve this target return with high accuracy, and even to generalize to returns that were not seen during training.

\textbf{Contributions:}
In this work, we experimentally show that MAP-Elites-based algorithms have considerable difficulty in producing consistent policies in uncertain domains.
We introduce a new metric, the policy \textit{spread}, which measures the consistency of policies in the BD space and propose a variant of MAP-Elites, called MAP-Elites Low-Spread (ME-LS), which drives the search process towards consistent policies by selecting them for their higher fitness and lower spread. We then introduce the Quality-Diversity Transformer (QDT), a behavior-conditioned model inspired by the DT that learns to achieve behaviors on demand and leverages supervised learning on datasets of trajectories generated by repertoires of policies.

We run experiments in simulated robotic environments based on the physics engine Brax~\cite{freeman2021brax} and show experimentally that: 1. ME-LS produces repertoires of consistent policies that replicate their behavior descriptors (BDs) over multiple episodes with varying initial conditions,  
2. The QDT compresses a whole repertoire into a single policy and demonstrates even better accuracy than ME-LS policies, 3. Finally, the QDT is capable, to a certain extent, to generalize to unseen behaviors.
All the code in this work
was created using the framework QDax~\cite{lim2022accelerated,chalumeau2023qdax}.

\section{Background}

\subsection{MAP-Elites}
MAP-Elites (ME)~\citep{mouret2015illuminating} is a simple yet efficient method for Quality-Diversity that has been successfully utilized to tackle a broad range of complex tasks including restoring function to damaged robots~\citep{cully2015robots}, optimizing self-assembling micro-robots~\citep{cazenille2019exploring} and designing games ~\citep{alvarez2019empowering}. The algorithm involves discretizing the behavior descriptor space, $\mathcal{B}$, into a repertoire of cells, known as niches, with the goal of populating each niche with an optimal solution.
An extension to the MAP-Elites algorithm, referred to as CVT MAP-Elites~\cite{DBLP:journals/corr/VassiliadesCM16}, employs Centroidal Voronoi Tesselations (CVT) to initially segment the repertoire into the intended number of cells.
Full pseudocode is available in Appendix~\ref{sec:me_pseudocode}.

MAP-Elites starts with an empty repertoire and a random set of initial solutions that are evaluated and placed into the repertoire according to a simple insertion criteria. If the cell corresponding to a solution's behavior descriptor is unoccupied, the solution is inserted into that cell. If there is already a solution in this cell, the new solution will only replace it if it possesses superior fitness.
At every iteration, a set of existing solutions are randomly selected from the repertoire and mutated to generate new solutions. These new solutions are then evaluated and inserted into the repertoire using the same insertion criteria as before. This cycle is repeated until either convergence is reached or a predetermined number of iterations is completed. MAP-Elites is a powerful and compelling method, however it has the drawback of producing solutions that are subject to high behavior and fitness variability in uncertain domains such as continuous control environments~\cite{flageat2022empirical}.

\subsection{The Decision Transformer}
The Transformer~\cite{vaswani2017attention} is a popular model architecture that was specifically designed for natural language processing (NLP) tasks such as language translation~\cite{devlin2018bert}, but has since been applied to a variety of other tasks, including image and speech recognition ~\cite{dong2018speech, dosovitskiy2020image}, text generation ~\cite{brown2020language} and sequence modeling for RL ~\cite{chen2021decision}. The key innovation of the Transformer is its use of self-attention mechanism, which allows the model to efficiently weigh the importance of different parts of the input when making predictions, rather than relying solely on the order of the input as in previous architectures.

Authors of the Decision Transformer (DT)~\cite{chen2021decision} propose a new approach to sequential decision making problems formalized as reinforcement learning using the Transformer architecture. They train the DT on collected experience (datasets of trajectories) using a sequence modeling objective --allowing for a more direct and effective way of performing credit assignment-- and show that it matches or exceeds the performance of state-of-the-art model-free offline RL algorithms.
They use a trajectory representation that allows a causal Transformer based on the GPT architecture~\cite{radford2018improving} to autoregressively model trajectories using a causal self-attention mask. Put simply, the DT predicts actions by paying attention to all prior elements in a trajectory. With this work, the authors bridge the gap between sequence modeling and reinforcement learning, proving that sequence modeling can serve as a robust algorithmic paradigm for sequential decision making.

\subsection{Related Work}
Several works have investigated the behavior and fitness estimation problem in uncertain domains. 
Authors of~\cite{cully2018hierarchical} aggregate multiple runs to evaluate a single solution. They perform insertion in the repertoire based on the average fitness over all episodes and based on the geometric median of all obtained BDs. Following the terminology introduced in~\cite{flageat2022empirical}, we name this method MAP-Elites-sampling and remark that, without having been named explicitly, a very similar approach has been used in~\cite{engebraaten2020framework}. We argue that MAP-Elites-sampling suffers from two issues: First, its focus is on finding better approximation of the true characteristics of solutions, rather than directly searching for more robust ones. Second, using the geometric mean over all BDs can result in a situation where a solution is qualified by a BD that it never actually achieves, which could prevent another, more adequate solution to be inserted in the repertoire. To support these claims, we include ME-Sampling as a baseline in our experiments. Deep-Grid~\cite{flageat2020fast} is a variant of MAP-Elites that employs an archive of similar previously encountered solutions to estimate the characteristics of a solution. However, \cite{flageat2022empirical} shows that Deep-Grid fails to find reproducible solutions as efficiently as MAP-Elites-sampling, and hypothesizes that because the method uses neighbours in the BD-space to approximate the true characteristics of solutions, it does not perform well in high dimensional search spaces --which are typical in neuroevolution tasks-- where the complex relation between genotypes and BDs can be confusing for this neighbourhood-based mechanism.
Adaptive-Sampling \cite{justesen2019map} is a method that discards solutions that are evaluated too many times outside of their main cells to keep only the most reproducible solutions. Authors of \cite{flageat2023uncertain} propose an extensive study of the reproducibility problem, they compare existing methods (including MAP-Elites Sampling, Deep-Grid and Adaptive-Sampling) and introduce new variations.
Still, as opposed to ME-LS, all these methods do not act directly on insertion but progressively measure the reproducibility of solutions and eliminate the least reproducible ones.
Policy Gradient Assisted MAP-Elites (PGAME)~\cite{nilsson2021policy} is a state-of-the-art QD algorithm which bridges the gap between evolutionary and policy gradient (PG) methods by introducing a PG variation operator in ME. In this work, we replicate all ME-based experiments and adapt them to PGAME to show that the stated problem and our conclusions are still valid in the presence of PG variations. 

In line with the Decision Transformer, the Trajectory Transformer~\cite{janner2021offline} achieves goal-conditioned trajectory generation in 2D MiniGrid environments, and authors of Multi-Game Decision Transformers~\cite{lee2022multigame} train a single model that plays up to $46$ Atari games with near-human performance. Even more impressively, GATO~\cite{reed2022generalist} is a generalist Transformer-based agent which tackles a wide variety of tasks including, among others, real and simulated robotic manipulations. A point of divergence between these works and ours is the tokenization scheme used: while they tokenize each dimension of each element given as input to their models, we follow the DT scheme and directly feed the continuous observations, actions and conditioning BD to the QDT. This has the advantage of drastically reducing the computing time and resources required as the model itself is smaller and does not grow as a function of the input dimension. Finally, \cite{jegorova2020behavioral} introduces a behavior-conditioned generative model which generates parameters of policies that are conditioned to achieve a target behavior. However, their model is not suited for deep neuroevolution and focuses on generating simple policies consisting of dozens of parameters.

\section{Problem Statement}

We consider sequential decision-making problems expressed as Markov Decision Processes (MDPs) defined as $\left(\mathcal{S}, \mathcal{A}, \mathcal{R}, \mathcal{T} \right)$ where $\mathcal{S}$ is the state space, $\mathcal{A}$ the action space, $\mathcal{R} : \mathcal{S} \times \mathcal{A} \rightarrow \mathbb{R}$ the reward function and $\mathcal{T}: \mathcal{S} \times \mathcal{A} \rightarrow \mathcal{S}$  the transition function.
We assume that both $\mathcal{S}$ and $\mathcal{A}$ are continuous and that both $\mathcal{T}$ and $\mathcal{R}$ are deterministic functions. Policies $\pi_{\theta}: \mathcal{S} \rightarrow \mathcal{A}$ are assumed to be implemented by neural networks parameterized by $\mathbf{\theta} \in \Theta$, which are called {\em solutions} to the problem. We denote by $\tau_{[\pi_{\theta}, s]} \in \Omega$ a trajectory of the policy $\pi_{\theta}$ starting from the initial state $s$.

Contrary to previous related works~\cite{pierrot2022diversity, pierrot2022multi, Nilsson2021, chalumeau2022neuroevolution} and due to the fact that we exclusively consider uncertain continuous control environments where the initial state is randomly sampled, we do not expect a direct mapping from a solution $\theta$ to its fitness nor to its behavior descriptor and rather consider fitnesses and behavior descriptors of {\em trajectories} played by $\pi_{\theta}$. 
The fitness function $F: \Omega \rightarrow \mathbb{R}$ takes a trajectory $\tau_{[\pi_{\theta} , s]}$ as input and measures its performance, defined as the sum of rewards obtained by policy $\pi_{\theta}$ during an episode starting from initial state $s$. It is used to estimate the actual fitness of the solution $\theta$, which can be theoretically defined as the expected value of the fitness for a given initial state distribution. We also introduce a behavior descriptor space $\mathcal{B}$ and a behavior descriptor extraction function $\mathbf{\xi}: \Omega \rightarrow \mathcal{B}$ that maps a trajectory $\tau_{[\pi_{\theta}, s]}$ to its behavior descriptor $\mathbf{\xi}(\tau_{[\pi_{\theta}, s]})$. Assuming that $\text{dist}_{\mathcal{B}}$ is a distance metric over $\mathcal{B}$, we define the \emph{spread} of $K$ trajectories as the mean distance between all pairs of behavior descriptors. We use it to estimate the spread of the solution $\theta$ (cf. Equation~\ref{eq:spread_computation}), which again can be theoretically defined as the expected value of the distance between the behavior descriptors of two trajectories for a given initial state distribution.
In most cases, we could also use the standard distance deviation or dispersion, defined as the mean distance to the mean behavior descriptor, which has better computational complexity, but requires the definition of the mean behavior descriptor, which is trivial when they belong to a vector space, but can be difficult in the general case. Furthermore, the difference in computational complexity has a negligible impact on our method, as we always consider a limited number of evaluations for each policy.

\begin{equation}
\label{eq:spread_computation}
    \psi \left(\mathbf{\theta}\right) \mathrel{\hat=} \underset{1\leq i<j\leq K}{\avg} \text{dist}_{\mathcal{B}}\left(\mathbf{\xi}(\tau_{[\pi_{\theta}, s_i]}),\mathbf{\xi}(\tau_{[\pi_{\theta}, s_j]})\right),
\end{equation}





\noindent where the states $\{s_i\}_{i=1,\ldots ,K}$ are randomly initialized. 

Put simply, the spread of a solution measures its tendency to obtain different behavior descriptors when playing multiple episodes with varying initial states. The lower the spread, the more consistent the policy is in the BD space. Note that in the experiments presented in this paper, $\mathcal{B}$ is a Euclidean space and we use the Euclidean distance over $\mathcal{B}$ as $\text{dist}_{\mathcal{B}}$.

\subsection{Environments}
The set of tasks we study is based on the
Brax suite~\cite{freeman2021brax}, a physics engine for large scale rigid body simulation written in Jax~\cite{jax2018github}. It is worth mentioning that all policies used in this work (including the QDT and policies produced by QD algorithms) are deterministic, thus the stochasticity is only the result of the variability in the initial states, which are randomly sampled from a Gaussian distribution.

\begin{figure}[ht]
\centering
    \includegraphics[width=0.4\textwidth]{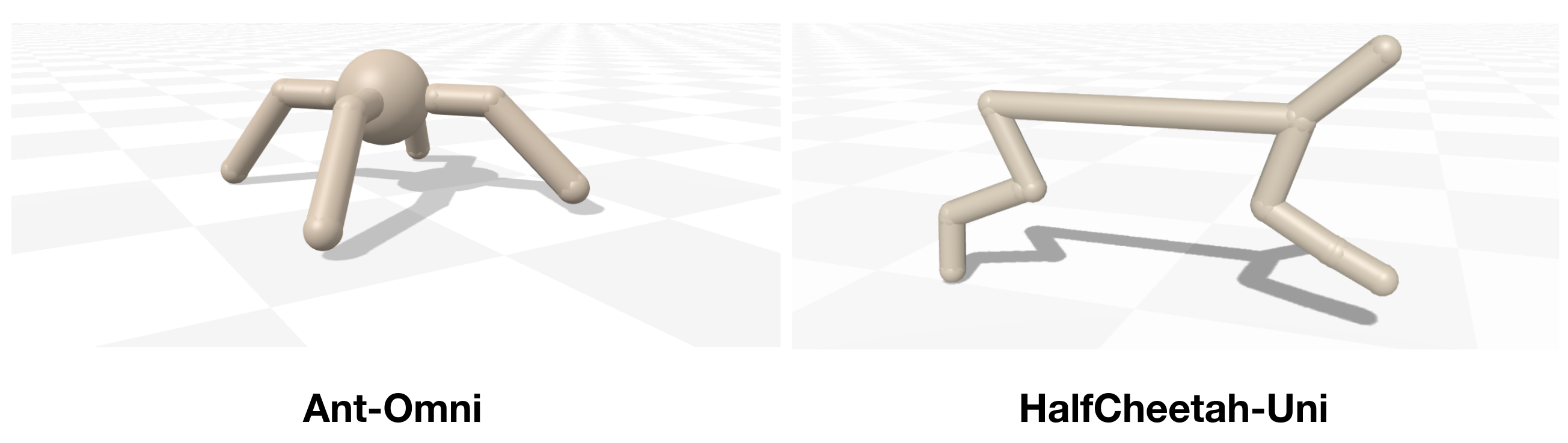}
    \caption{Illustration of the benchmark tasks.
    }
    \label{fig:environments}
\end{figure}
\begin{figure*}[ht]
\centering
    \includegraphics[width=1\textwidth]{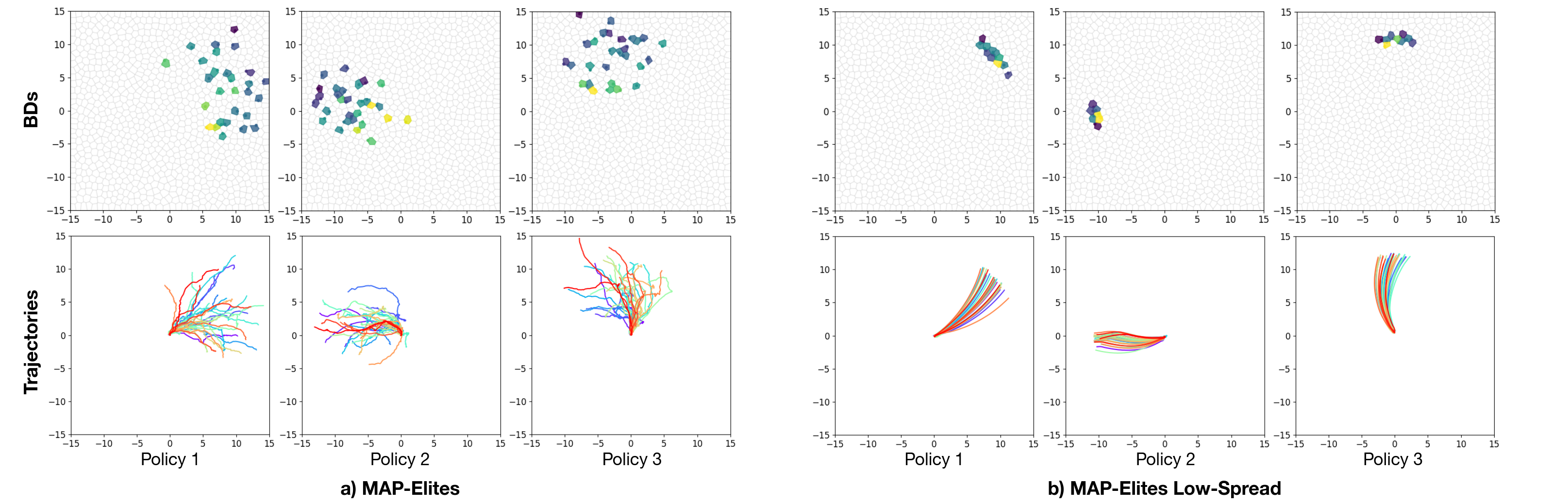}
    \caption{Illustration of the reproducibility problem in Ant-Omni. We select three representative policies from final repertoires that have been generated by a) MAP-Elites and b) MAP-Elites Low-Spread, our proposed variant, and play 30 episodes with each policy using slightly varying initial states. The top row depicts the final BDs obtained by each policy and the bottom row represents the corresponding entire trajectories in the behavior space. Each color represents a different random seed (initial condition).}
    \label{fig:qd_problem}
\end{figure*}

We evaluate the capacity of our method to perform in continuous control locomotion tasks that feature substantially distinct behavior spaces, as well as high dimensional observation and action spaces. In these tasks, the challenge is to move legged robots by applying torques to their joints via actuators. These types of environments are particularly challenging for evolutionary algorithms as they typically necessitate a significant number of interactions with the environment to develop high-performing policies. We follow the terminology introduced in~\cite{chalumeau2022neuroevolution} to name environments and distinguish between omnidirectional and unidirectional environments by providing them with distinct names. All episodes in both environments have a maximum length of $250$ time steps.

\textbf{Ant-Omni} is an exploration-oriented environment in which a four-legged ant robot must move on the 2D plane while minimizing the control energy~\cite{flageat2022benchmarking}. This environment is similar to the popular Ant-Uni environment~\cite{chalumeau2022neuroevolution} (as it involves the same articulated ant) but instead of trying to move as fast as possible in a single direction, the goal is to reach any location on the surface.
In this environment, the BD space is defined as the 2D plane and behavior descriptors are 2-dimensional vectors computed as the $x$ and $y$ positions of a solution at the end of an episode. We chose to restrict the BD space of Ant-Omni to $[-15,15]$ on both axes as methods presented in this work tend to produce solutions within this range. 
The reward signal is defined as the negative energy consumption at each time step and simply ensures that policies are constrained to produce energy-efficient behaviors. It is worth noting that this environment is primarily intended to evaluate the exploration abilities of algorithms, as the reward is relatively easy to optimize \textemdash an optimal policy would simply remain static. BD extraction function and fitness function for the Ant-Omni environment are defined as:
\begin{equation}
\label{eq:ant_omni_def}
    \begin{cases}
    \xi(\tau)_{\textit{Ant-Omni}} = (x_T, y_T)\\
    F(\tau)_{\textit{Ant-Omni}} = - \sum\limits_{t=1}^{T}{||\textbf{a}_t||_2}\\
    \end{cases}
\end{equation}
where $T$ is the number of transitions in the trajectory, $||.||_2$ is the Euclidean norm, $\textbf{a}_t$ is the action vector that corresponds to torques applied to the robot joints, and $x_T$ and $y_T$ the positions of the robot's center of gravity on both axes at the end of the trajectory.


\textbf{Halfcheetah-Uni} is a popular benchmark environment in which the agent must run as fast as possible in the forward direction while maximizing a trade-off between speed and energy consumption. In this environment, the BD space is defined as all the possible patterns of movement \textemdash or gaits\textemdash the bipedal agent can use to run. Behavior descriptors are defined as the proportion of time each foot of the agent is in contact with the ground. This definition is commonly used in other related works for tasks of similar nature~\cite{pierrot2022multi, colas2020scaling, cully2015robots}.
The reward signal is defined as the forward distance covered between each time step minus a penalty for energy consumption.
BD extraction function and fitness function for the Halfcheetah-Uni environment are defined as:
\begin{equation}
\label{eq:halfcheetah_uni_def}
    \begin{cases}
    \xi(\tau)_{\textit{Halfcheetah-Uni}} = (\frac{1}{T} \sum\limits_{t=1}^{T}{c_{i,t}})_{i=1,2}\\
    F(\tau)_{\textit{Halfcheetah-Uni}} = \sum\limits_{t=1}^{T}{\frac{x_t-x_{t-1}}{\Delta_t}} - \sum\limits_{t=1}^{T}{||\textbf{a}_t||_2}\\
    \end{cases}
\end{equation}
where $x_t$ is the position of the robot's center of gravity at time $t$ along the forward axis and $\Delta_t$ the time step, and $c_{i,t} = 1$ if leg $i$ is in contact with the ground at time $t$ and $0$ otherwise. 

\subsection{The Reproducibility Problem}
\label{sec:reproductibility_problem}

Uncertain domains such as continuous control environments are known to be challenging for evolutionary methods. Apart from the fact that these methods are usually less efficient when facing high-dimensional search spaces~\citep{colas2020scaling}, they also tend to generate policies that exhibit a high degree of variability in their behaviors and performances~\cite{flageat2020fast, flageat2022empirical}. As a result, repertoires of solutions generated by QD algorithms have limited re-usability in the sense that the solutions they store rarely replicate the behaviors and performances for which they were retained. Even though the dynamics of the environment and the policy themselves may be deterministic, an ideal policy should be robust to varying initial conditions and demonstrate consistent behaviors and performances, particularly in the context of simulated robotics where the transfer to real applications is 
dependent on the robustness of such policies.

Figure~\ref{fig:qd_problem} illustrates this problem for MAP-Elites in the Ant-Omni environment. 
The reproducibility problem is actually twofold: first, it appears that the policies from the MAP-Elites repertoire exhibit a very high spread in the behavior space (as defined by Equation~\ref{eq:spread_computation}), meaning that a policy is unable to reproduce the results that were used to insert it in the repertoire. Table~\ref{tab:dist_spread_table} contains average spread values for MAP-Elites and PGA-MAP-Elites in both environments. We argue that this pitfall of MAP-Elites based algorithms comes from the single evaluation scheme being used (see the MAP-Elites pseudocode in Appendix~\ref{sec:me_pseudocode}), which drives the search process towards solutions that show high variability and that have been lucky during their single evaluation episode. Second, it appears that not only the MAP-Elites policies display a high spread in the behavior space, but they also produce inconsistent and irregular trajectories as depicted in the bottom row of Figure~\ref{fig:qd_problem}a). We refer to these trajectories as being irregular because the ant robot does not move steadily from the starting point to its final position, but rather follows a shaky trajectory that often changes course. 

Similar analyses for the Halfcheetah-Uni environment can be found in Appendix~\ref{sec:add_results_qd_problem} where we additionally show that PGA-MAP-Elites suffers from the same shortcomings. 
In the next section, we propose a new approach that augments MAP-Elites based algorithms with an additional insertion criterion based on the spread computation and show that it solves both of the above mentioned problems. We later show that the QDT alone is able to mitigate these problems but also that the supervised learning phase of the QDT largely benefits from trajectories that have been generated by steady and consistent policies.

\makeatletter
\makeatother

\SetKwComment{Comment}{/* }{ */}

\SetArgSty{textnormal}

\begin{algorithm}
    \small
    \SetAlgoLined
    \DontPrintSemicolon
    \SetKwInput{KwInput}{Given}
    \KwInput{
    \begin{itemize}
        \item Max iteration number $I$
        \item Number of initialization solutions $G$
        \item Number of evaluations per solution $E$
        \item MAP-Elites repertoire $\mathbb{M}$
    \end{itemize}
    }
    \texttt{\\}
    \tcp{Main loop}
    $iteration\_number \xleftarrow{} 0$\;
    \While{$iteration\_number < I$}{
    
        \texttt{\\}
        \tcp{Initialize by generating G random solutions}
         \textbf{if} $iteration\_number < G$ \textbf{then}\;
          \Indp  $\mathbf{x'} \xleftarrow{} random\_solution()$
        
        \Indm \tcp{Sampling and mutation}
        \textbf{else}\;
         \Indp $\mathbf{x} \xleftarrow{} random\_selection(\mathbb{M})$\;
               $\mathbf{x'} \xleftarrow{} random\_genetic\_mutation(\mathbf{x})$\;

        \Indm \tcp{Evaluation}
        $evaluate(\mathbf{x'})$ over $E$ trajectories\;
        Compute fitness of $\mathbf{x'}$ as its avg. fitness over $E$ trajectories\;
        Compute BD of $\mathbf{x'}$ as its most frequent BD over $E$ trajectories\;
        Compute the spread of $\mathbf{x'}$ in the BD space as given by Eq.~\ref{eq:spread_computation}

        \tcp{Insertion in repertoire}
        Insert $\mathbf{x'}$ in $\mathbb{M}$ only if its fitness is higher and its spread is lower than the corresponding solution already in $\mathbb{M}$\;
        
        $iteration\_number = iteration\_number + 1$
    }

    \caption{MAP-Elites Low-Spread}
    \label{alg:ME-LS}
\end{algorithm}

\begin{figure}[ht]
\centering
    \includegraphics[width=0.4\textwidth]{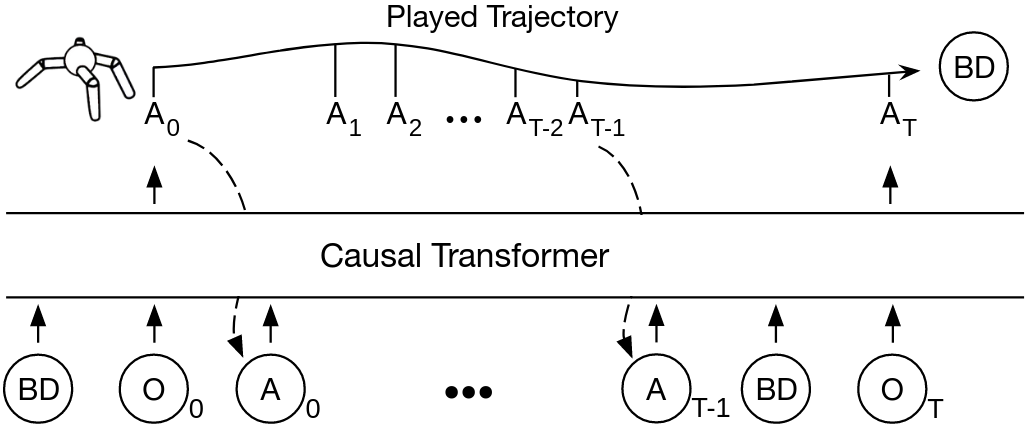}
    \caption{The QDT autoregressively playing an evaluation episode by conditioning on a target BD. At any given time step $t$, the QDT generates an action by looking at all elements of the trajectory that precede $t$. The first action $A_0$ is generated given the target BD and the first observation $O_0$, while the last action $A_T$ is generated given the whole trajectory of target BDs, observations and previously generated actions. Note that the target BD is the same for the entire trajectory.}
    \label{fig:qdt}
\end{figure}
\vspace{-0.4cm}

\section{Methods}
\subsection{MAP-Elites Low-Spread}

We introduce the MAP-Elites Low-Spread (ME-LS) algorithm, our variant of the original MAP-Elites algorithm (ME)~\cite{mouret2015illuminating} that thrives the search process towards solutions that are consistent in the behavior space for uncertain domains. The full pseudocode is presented in Algorithm~\ref{alg:ME-LS}.
ME-LS uses the global structure of ME except for two aspects. First, solutions are evaluated over multiple episodes ($N=10$) and second, solutions are inserted into the repertoire if they prove to have a higher fitness \textit{and} a lower spread than the solutions already contained.
More precisely, the overall operation of ME-LS can be described in 3 principal steps: 1. ME-LS create new solutions through mutations, 2. It evaluates them over multiple episodes and compute their average fitnesses and their most frequent BDs\footnote{Since the behavior space is continuous, we consider that two trajectories have the same behavior descriptor if they both belong to the same cell in the repertoire.}, 3. These new solutions are inserted in the repertoire if they have better fitness and lower spread than the already stored solutions, or if the corresponding cells are empty.

As shown in Figure~\ref{fig:qd_problem}b), the policies generated by ME-LS exhibit highly consistent BDs over 30 episodes and regular, steady trajectories in the behavior space. It is clear that the additional insertion criterion forces solutions to be consistent in the behavior space, and we hypothesize that this additional constraint indirectly forces the selection process towards solutions that produce smooth, regular trajectories. We also show that similar conclusions hold for the PGA-ME algorithm in Appendix~\ref{sec:add_results_qd_problem}, and detail its Low-Spread version (PGAME-LS) in Algorithm~\ref{alg:PGAME-LS}. Table~\ref{tab:dist_spread_table} contains average spread values for all methods in both environments. It is important to note that simply performing multiple evaluations to better characterize a solution, as in ME-Sampling, does not help to increase its consistency in the behavior space. To solve this we argue that it is preferable to optimize directly for this purpose. 

Finally, we present results that compare ME, ME-LS, PGAME and PGAME-LS in Appendix~\ref{sec:add_results_qd_algos}.
Although the Low-Spread versions require several evaluations per solution, their convergence rate --in terms of number of interactions with the environment-- is in fact similar to that of original methods, resulting in a comparable sample efficiency.
Unsurprisingly however, ME-LS and PGAME-LS present significantly worse training metrics (coverage, max fitness and QD score) than their original counterparts, due to the fact that they prevent the insertion of lucky solutions in the repertoire. We conduct a reassessment experiment in Appendix~\ref{sec:reassessment_xp} where we re-evaluate final repertoires produced by all methods and show that this performance gap in training metrics is not representative of the true quality of the final repertoires. Results of this experiment are reported in Table~\ref{tab:metrics_xp} and prove that after re-evaluation, ME-LS and PGAME-LS repertoire obtain better coverages and QD Scores, and comparable maximum fitnesses to ME and PGAME.

\begin{figure}[ht]
\centering
    \includegraphics[width=0.5\textwidth]{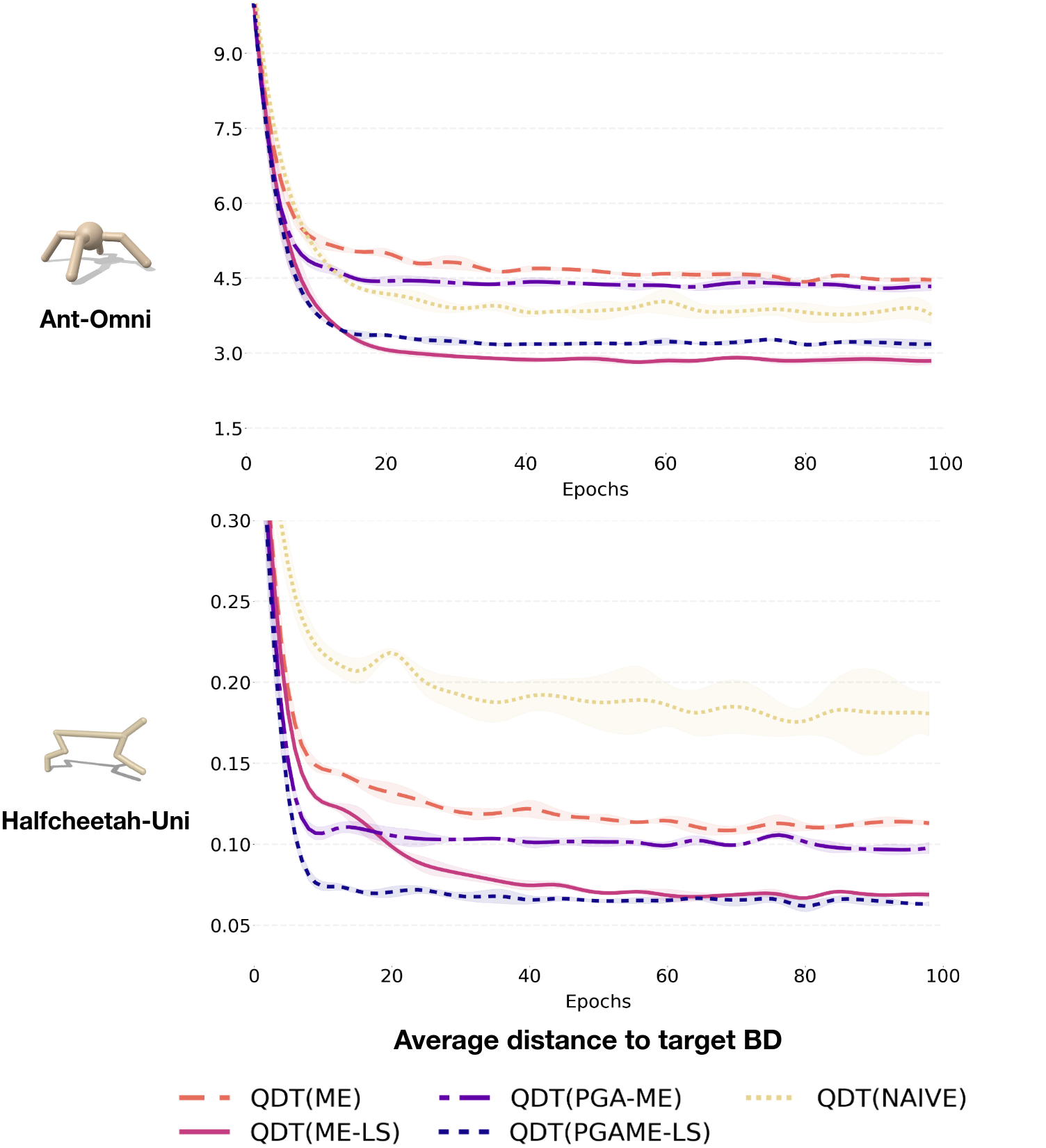}
    \caption{Results of evaluations of the QDT through the training process (average values and std ranges on 3 seeds). We evaluate the model over multiple goals (target BDs) which cover the behavior space and report the total average Euclidian distance to these goals. The QDTs that trained on datasets created by Low-Spread methods, whether using ME-LS or PGAME-LS, show significantly better performances.}
    \label{fig:qdt_training}
    \vspace{-0.5cm}
\end{figure}

\begin{figure*}[ht]
\centering
    \includegraphics[width=0.94\textwidth]{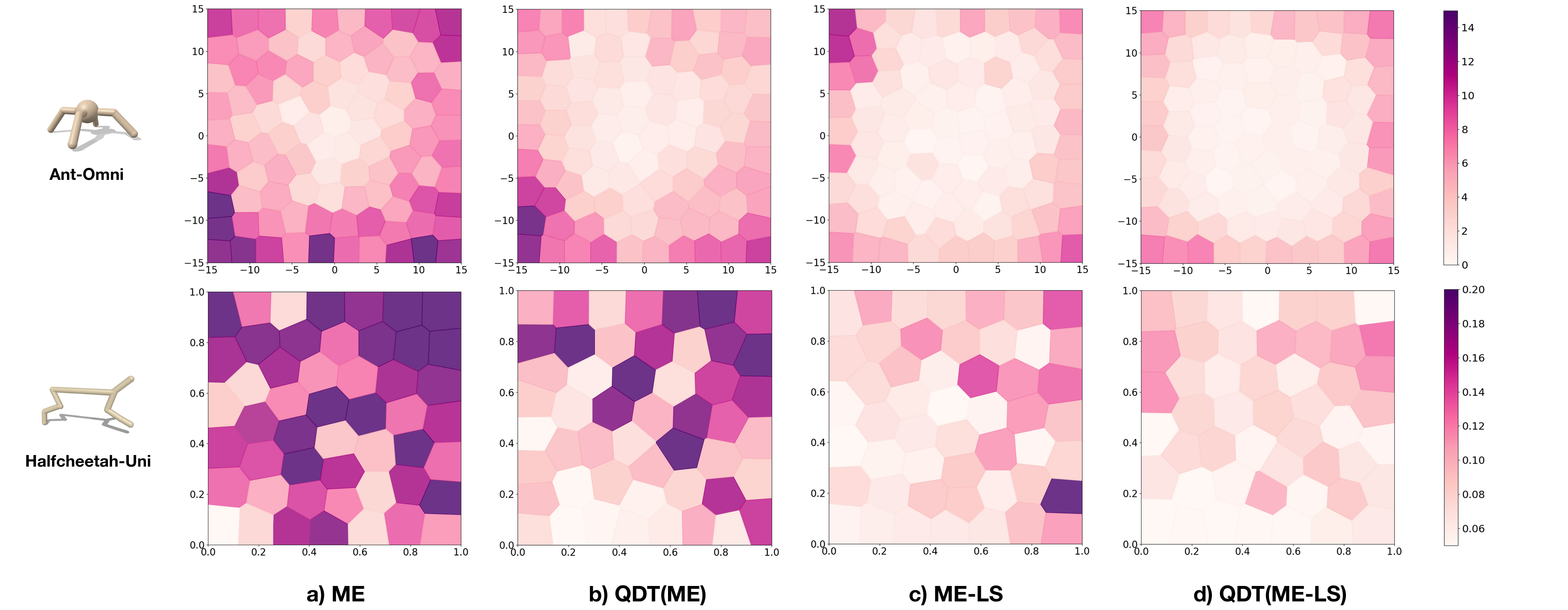}
    \caption{Results of the accuracy experiment. This experiment can be described in 2 steps: 1. We select multiple evaluation goals (target BDs) in the behavior space, 100 and 50 for Ant-Omni and Halfcheetah-Uni respectively. To get meaningful goals, we simply compute a CVT of the BD space in which goals are the centers of each zone, 2. For each goal, we play 10 episodes and plot their average Euclidean distance to the goal. For ME and ME-LS, trajectories are played by the nearest policy to the goal in the repertoire. For the QDT, we simply condition it on the goal. Distance is represented by color: lighter is better. The QDT(ME-LS) appears to be the most accurate method to achieve behaviors on demand.}
    \label{fig:results_precision}
    \vspace{-0.0cm}
\end{figure*}

\subsection{The Quality-Diversity Transformer}
This section introduces the Quality-Diversity Transformer (QDT), a model inspired by the Decision Transformer~\cite{chen2021decision} that autoregressively models trajectories and produces behaviors on-demand by conditioning on a target behavior descriptor, as shown in Figure~\ref{fig:qdt}.
Although very similar, the Quality-Diversity Transformer differs from the DT in that it conditions decisions on a target behavior rather than on the return to be obtained during an episode.

A key aspect of the model is the trajectory representation used as input: $\tau = (BD, O_0, A_0, BD, O_1, A_1, ... , BD, O_T, A_T)$. This trajectory structure enables the model to learn useful patterns and to conditionally generate actions at evaluation time. It should be noted that, contrary to the Decision Transformer, we simply use the same conditioning BD at each time step and do not use a representation analogous to the return-to-go, which is a dynamic conditioning introduced in~\cite{chen2021decision} to capture the return to be achieved until the end of the episode at any time step. We tested it and obtained poorer results. We also tested a version of the QDT where the conditioning BD appears only once at the beginning of the episode, the intuition being that the attention mechanism of the Transformer should be able to focus on this element, even if it appears only once. Results were very similar to those presented in this paper but marginally inferior, hence the choice of preferring this representation.
We feed trajectories of length $3T$ to the QDT, as we have $3$ elements at each time step (one for each modality: conditioning BD, observation and action). Note that the QDT takes raw continuous inputs from the environment which are not normalized. We compute embeddings for these elements by learning a linear layer for each modality, which projects elements to the embedding dimension, followed by layer
normalization~\cite{ba2016layer}. Time step embeddings are learned for each of the $T$ time steps and added to their corresponding elements. This differs from traditional positional embeddings used in Transformers as one time step embedding corresponds to three elements. Finally, the trajectory is processed by the Transformer model which predicts one continuous action vector for each time step.

During training, we use a dataset of offline trajectories generated by policies from a QD repertoire and leverage supervised learning to train the QDT over entire trajectories from the dataset, which is a point of divergence with the Decision Transformer as they use randomized reduced windows instead of whole trajectories. The prediction head corresponding to the input token $O_t$ is trained to predict $A_t$ with mean-squarred error loss as we run our experiments on continuous action spaces (see Figure~\ref{fig:teaser} part 3: "Supervised Learning"). Because of the trajectory structure and since the model is a causal Transformer (GPT-2 based~\cite{radford2019language}) that can only attend to previous inputs at any given time step, we can make forward passes over minibatches of trajectories and compute the loss for all time steps at once.
An exhaustive description of the QDT training process can be found in Appendix~\ref{sec:qdt_pseudocode}.

To evaluate the QDT, we simply condition it on a target behavior descriptor (BD) and feed it with the first observation as given by the environment. The model generates the first action, which is played next in the environment and appended to the trajectory of inputs for the next inference. We unroll a whole episode in the environment following this procedure and measure the QDT's performance by computing the Euclidean distance between the conditioning BD and the BD actually achieved by the model during the episode.

\subsection{Dataset Creation}
\label{sec:dataset_creation}

Dataset generation is the second stage of our full pipeline illustrated in Figure~\ref{fig:teaser}. For this purpose, we divide the repertoire into large zones (50 for Halfcheetah-Uni, 100 for Ant-Omni) and select the best policy for each zone. To do so, each candidate policy of a given zone plays a few evaluation episodes and the policy that most often produces BDs corresponding to the zone is selected. We thus obtain a total of 50 (resp. 100) selected policies, and make them play trajectories that are stored into a dataset. Datasets for both environments are constituted of $300,000$ trajectories, or equivalently $75$ million transitions, given that each trajectory consists of 250 time steps.

\begin{figure*}[ht]
\centering
    \includegraphics[width=1\textwidth]{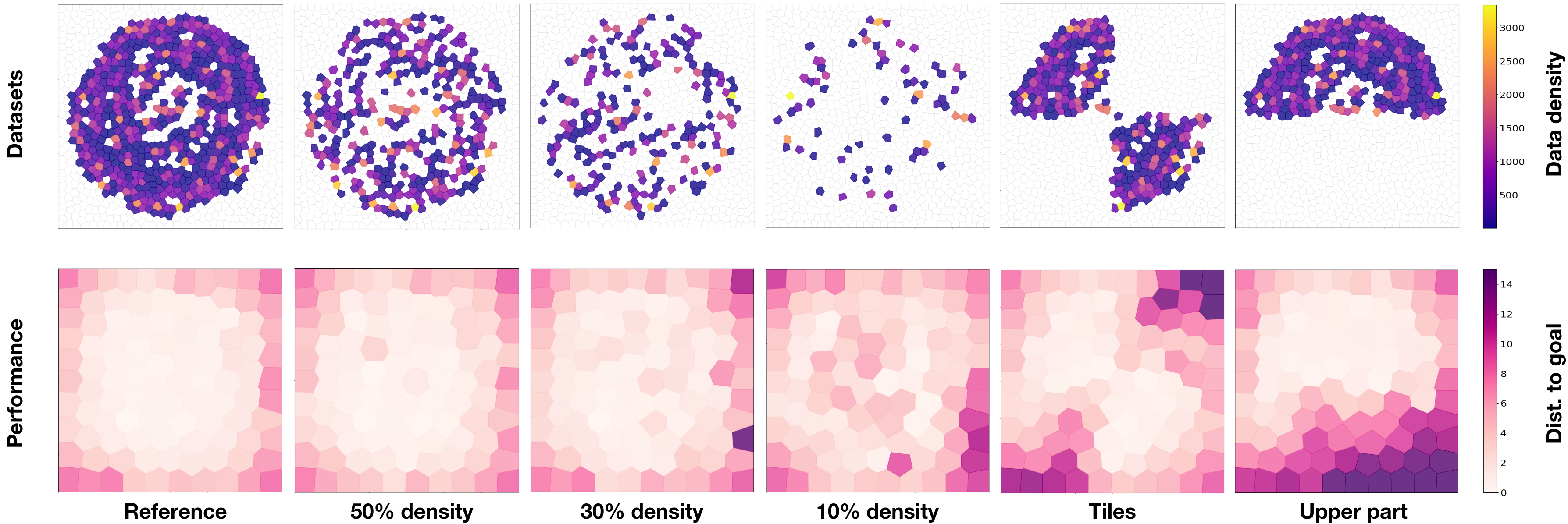}
    \caption{Results of the QDT generalization experiment in Ant-Omni. In this experiment we run accuracy experiments (bottom row) on truncated datasets (top row) which are deprived of a part of their trajectories. The QDT shows strong interpolation ability on the 50\%, 30\% and 10\% density datasets and a more limited ability to extrapolate in "Tiles" and "Upper part" datasets where entire zones of the BD space are deprived of data.}
    \label{fig:results_generalization}
\end{figure*}
\vspace{-0.2cm}
\section{Results}

\subsection{QDT Training and Ablations}
\label{sec:training_and_ablations}
During the supervised learning process, we periodically evaluate the QDT and report its average Euclidean distance to target BDs. For each evaluation phase, these target BDs --or goals-- are chosen to be representative of the whole behavior space, as we want to assess the model's capacity to reach all zones of the space with high accuracy. To do so, we compute a Centroidal Voronoi Tesselation of the BD space and use each centroid as a conditioning goal for the QDT. We use $50$ and $100$ evaluation goals for Halfcheetah-Uni and Ant-Omni respectively, although these values are arbitrary, we consider that they allow a fair coverage of the BD space in both cases. 
We run multiple ($N=10$) episodes for each goal and compute the average distance per goal --which can be visualized as shown in Figure~\ref{fig:results_precision}-- as well as the overall average distance for all goals.

Figure~\ref{fig:qdt_training} plots the overall average distance in both environments for different models through the training process. In both tasks, we train the models for $100$ epochs and perform evaluation every $5$ epochs. Models are named after the QD algorithm that was used to generate the dataset: QDT(ME-LS) refers to the QDT that is trained on a dataset generated by MAP-Elites Low-Spread policies, which constitutes our full algorithm as depicted in Figure~\ref{fig:teaser}. QDT(ME) is an ablation which refers to the QDT that is trained on a dataset generated by MAP-Elites policies, and so on for PGAME and PGAME-LS. The QDT(Naive) model corresponds to an ablation where we do not apply the dataset creation method described in Section~\ref{sec:dataset_creation} and simply run all policies from a ME-LS repertoire to generate the dataset. Results show that the trend is the same among the two algorithm families (ME and PGAME): Low-Spread based QDTs outperform their vanilla counterparts and achieve significantly lower average distance to target BDs. We argue that this is the direct result of using more consistent policies to create the dataset, and we hypothesize that learning to replicate a skill that has been demonstrated in a steady and accurate manner is inherently easier than learning to replicate irregular demonstrations, as depicted in Figure~\ref{fig:add_qd_problem}. Lastly, it appears that the dataset creation method of Section~\ref{sec:dataset_creation} is crucial to the good performance of the model, whose QDT(Naive) ablation especially struggles in Halfcheetah-Uni.

\begin{table} 
    \small
    \caption{Results of the accuracy experiment (described in Figure~\ref{fig:results_precision}). For each environment and algorithm, we present the average distance over all goals (target BDs) and the overall average spread. Each experiment is repeated over 5 seeds and reported with average values and standard deviations.
    }
        \centering
        \label{tab:dist_spread_table}
        \begin{tabular}{ccccc}
            \toprule

            \textbf{Method} & \multicolumn{2}{c}{Ant-Omni}  & \multicolumn{2}{c}{Halfcheetah-Uni}\\
                & Avg. dist. & Spread & Avg. dist. & Spread\\
                & & & ($\times10^2$) & ($\times10^2$)\\
            \midrule
            ME & $ 6.38\pm .15$ & $ 3.76\pm .03$ &  $ 16.33\pm .47$ & $ 15.45\pm .45$\\
            PGAME & $6.39 \pm .14$ & $ 3.99\pm .09$ & $17.67\pm .48$ & $ 16.00\pm .81$\\
            ME-LS & $3.01\pm .05$ & $ 2.33\pm .01$ & $10.01\pm .47$ & $ 10.01\pm .45$\\
            PGAME-LS & $3.13\pm .04$ & $ 2.24\pm .04$ & $10.00\pm .00$ & $ 10.67\pm .50$\\
            ME-Sampling & $5.10\pm .19$ & $ 4.50\pm .11$ & $15.40\pm .90$ & $ 14.12\pm .95$\\
            \midrule
            QDT (ME) & $4.64\pm .10$ & $ 4.66\pm .07$ & $12.00\pm .00$ & $ 13.00\pm .00$\\
            QDT (PGAME) & $4.45\pm .05$ & $ 4.18\pm .16$ & $12.00\pm .00$ & $ 12.00\pm .00$\\
            QDT (ME-LS) & ${\bf 2.86\pm .07}$ & $ {\bf1.99\pm .09}$ & $9.09\pm .32$ & $ 9.20\pm .22$\\
            QDT (PGAME-LS) & $3.06\pm .03$ & $ 2.32\pm .06$ & ${\bf7.34\pm .42}$ & ${\bf8.40\pm .37}$\\
            \bottomrule
        \end{tabular}
        \vspace{-0.3cm}
\end{table}

\subsection{Accuracy Experiment}
\label{sec:accuracy_xp}

This experiment aims to answer the following question: \textbf{Which method is more accurate and consistent in achieving a target behavior?}
To answer this question, we test the ability of ME, ME-LS, QDT(ME) and QDT(ME-LS) to achieve target BDs on demand and measure their consistency and accuracy. Similar to the evaluation method described in Section~\ref{sec:training_and_ablations}, we choose evaluation goals that cover the behavior space
and run multiple evaluation episodes ($N=10$) for each goal.
For ME and ME-LS, we run the solution of the repertoire which is closest to each respective goal. For QDT models, we select them according to their best training epoch (see Figure~\ref{fig:qdt_training}) and condition them directly on the goals (target BDs).

Figure~\ref{fig:results_precision} depicts the average distance per goal for each method. Color indicates whether, on average over the 10 episodes, we were able to reach the aimed point in the BD space (lighter is better). Importantly, it should be noted that in Ant-Omni, the robot has a size of roughly 2 units of distance and navigates on a plane of 30 units of distance on both axes. We consider achieving a distance to the target BD $\le 2$ a good performance. Likewise, the BD space of Halfcheetah-Uni ranges from $0$ to $1$ on both axes and we consider a distance to the target BD $\le 0.1$ a good performance. In both environments, results show that ME solutions hardly achieve the targeted behaviors. The QDT(ME) significantly improves over ME policies but is still short of producing accurate BDs. We hypothesize that the Transformer model helps to generalize on the data produced by ME policies but is still hampered by their irregular trajectories (we discuss the generalization abilities of the QDT in the next section). Finally, ME-LS and QDT(ME-LS) both demonstrate high accuracy for almost all goals in both environments. Note that in Ant-Omni, all methods struggle to reach the most outer goals, which correspond to a zone where no policy --hence no data-- is available. We show that similar results and conlusions hold for PGAME and its variations in Appendix~\ref{sec:add_accuracy_xp}.

Table~\ref{tab:dist_spread_table} repeats this experiment on multiple seeds and presents the overall average distance to goals and the overall average spread for all methods. In line with the results reported in Figure~\ref{fig:results_precision}, QDTs trained on datasets created by ME-LS and PGAME-LS policies show the best ability to achieve target behaviors while QDT(ME) and QDT(PGAME) confirm their superiority over ME and PGAME respectively. Additionally, we observe that the ME-Sampling baseline significantly improves over ME but is still far from the performance of ME-LS, particularly in Halfcheetah-Uni where we hypothesize that the geometric mean BD computation qualifies solutions in BD cells that they never actually achieve.

\vspace{-0.2cm}
\subsection{Generalization Experiment}
\label{sec:generalization_experiment}
This experiment aims to understand the generalization abilities of the QDT. Importantly, we want to distinguish between interpolation, the model's ability to generalize between BDs existing in the dataset, and extrapolation, the model's ability to reach BDs that are beyond existing examples. To do so, we sparsify datasets generated by ME-LS policies by pruning trajectories using different pruning schemes and train QDTs on these truncated datasets. 

Figure~\ref{fig:results_generalization} presents accuracy results for the Ant-Omni task. The top row shows the different datasets used in this experiment, color depicts the trajectory density in the BD space, light color corresponds to a high density zone. The bottom row shows accuracy experiments (similar to Figure~\ref{fig:results_precision}) for the QDTs trained on the corresponding datasets. The 50\%, 30\% and 10\% density datasets aim to measure the interpolation capacity of the QDT, while the "Tiles" and "Upper part" datasets aim to measure its extrapolation capacity. Our model demonstrates strong interpolation capabilities and is able to achieve behaviors with reasonable accuracy even in the 30\% and 10\% density settings. However, its extrapolation capabilities appear more limited as can be seen in the "Tiles" and "Upper part" settings. Appendix~\ref{sec:add_generalization_experiment} present generalization results for Halfcheetah-Uni.

\section{Conclusion}
This work introduced MAP-Elites Low-Spread, a QD algorithm that allows neuroevolution of diverse and consistent solutions in uncertain domains, and the Quality-Diversity Transformer, a single policy that achieves target behaviors with high accuracy by using behavior-conditioning. We showed that the QDT benefits from steady trajectories generated by ME-LS policies and that it is the best option to achieve behaviors on demand while being able, to some extent, to generalize to unseen zones of the BD space. We believe that an interesting future direction would be to apply the QDT to non-uncertain domains where there is no need to couple it with a Low-Spread-based QD algorithm, and benefit from its generalization capacities.

\begin{acks}

This work was supported with Cloud TPUs from Google's TPU Research Cloud (TRC) and has received funding from the European Commission’s Horizon Europe Framework Program under grant agreement No 101070381 (PILLAR-robots project).
Work by Nicolas Perrin-Gilbert was partially supported by the French National Research Agency (ANR), Project ANR-18-CE33-0005 HUSKI. Research 

\end{acks}

\bibliographystyle{ACM-Reference-Format}
\bibliography{qdt}

\clearpage
\newpage

\appendix

\section{Algorithms Pseudocodes}
\label{sec:algo_pseudocodes}
All algorithms are implemented based on the QDax framework \cite{lim2022accelerated, chalumeau2023qdax}, which provides highly parallelized versions of many state-of-the-art QD and RL algorithms using the Brax physics engine \cite{freeman2021brax}, and allows to create custom algorithms by providing basic building blocks. The duration of a MAP-Elites (resp. PGA-MAP-Elites) run requiring hundreds of millions of interactions with the environment does not exceed a few hours on modern GPUs using QDax.

\subsection{MAP-Elites}
\label{sec:me_pseudocode}
Algorithm~\ref{alg:ME} depicts the standard MAP-Elites algorithm \cite{mouret2015illuminating}. Contrary to its Low-Spread version, MAP-Elites evaluates each solution only once to measure fitness and BD. While in practice, the main operations (selection, variation and insertion) are paralellized over batches of solutions at each iteration, Algorithm~\ref{alg:ME} portrays the canonical form originally introduced in \cite{mouret2015illuminating} for the sake of readability. In this work, we used a batch size of $1000$ for both environments to speed-up computation time.
\makeatletter
\makeatother

\SetKwComment{Comment}{/* }{ */}

\SetArgSty{textnormal}

\begin{algorithm}
    \small
    \SetAlgoLined
    \DontPrintSemicolon
    \SetKwInput{KwInput}{Given}
    \KwInput{
    \begin{itemize}
        \item Max iteration number $I$
        \item Number of initialization solutions $G$
        \item MAP-Elites repertoire $\mathbb{M}$
    \end{itemize}
    }
    \texttt{\\}
    \tcp{Main loop}
    $iteration\_number \xleftarrow{} 0$\;
    \While{$iteration\_number < I$}{
    
        \texttt{\\}
        \tcp{Initialize by generating G random solutions}
        \If{$iteration\_number < G$}{
            $\mathbf{x'} \xleftarrow{} random\_solution()$
        }
        \tcp{Sampling and mutation}
        \Else{
            $\mathbf{x} \xleftarrow{} random\_selection(\mathbb{M})$\;
            $\mathbf{x'} \xleftarrow{} random\_genetic\_mutation(\mathbf{x})$\;
        }
            
        \texttt{\\}
        \tcp{Evaluation}
        \For{$i = 0 \xrightarrow{} E$}{
            $evaluate(\mathbf{x'})$ over $1$ trajectory
        }
        Compute fitness and BD of $\mathbf{x'}$

        \texttt{\\}
        \tcp{Insertion in repertoire}
        Insert $\mathbf{x'}$ in $\mathbb{M}$ only if its fitness is higher than the corresponding solution already in $\mathbb{M}$\;

        \texttt{\\}
        $iteration\_number = iteration\_number + 1$
    }

    \caption{MAP-Elites}
    \label{alg:ME}
\end{algorithm}

\subsection{PGA-MAP-Elites}
\label{sec:pgame_pseudocode}
The Policy Gradient Assisted MAP-Elites (PGA-MAP-Elites or simply PGAME) \cite{nilsson2021policy} algorithm builds upon MAP-Elites and introduces a variation operator based on policy gradient reinforcement learning to optimize for fitness.
PGA-MAP-Elites uses a replay buffer to store the transitions experienced by policies from the population during evaluation steps. It also employs an actor-critic model, using the TD3 algorithm \cite{fujimoto2018addressing} to train on the stored transitions. The critic is utilized at each iteration to calculate the policy gradient estimate for half of the offsprings selected from the MAP-Elites repertoire. A separate actor, known as the greedy controller, is also trained with the critic. This greedy controller is updated at each iteration and added to the MAP-Elites repertoire. Unlike other individuals, the greedy controller is never discarded, even if its fitness is lower than other individuals with similar behavior. Algorithm~ \ref{alg:PGAME} contains the pseudocode for PGA-MAP-Elites.
\makeatletter
\makeatother

\SetKwComment{Comment}{/* }{ */}

\SetArgSty{textnormal}

\begin{algorithm}
    \small
    \SetAlgoLined
    \DontPrintSemicolon
    \SetKwInput{KwInput}{Given}
    \KwInput{
    \begin{itemize}
        \item Max iteration number $I$
        \item Sample size $N$
        \item MAP-Elites repertoire $\mathbb{M}$
        \item Replay Buffer $\mathbb{B}$
        \item A critic network $Q_v$
    \end{itemize}
    }
    
    \texttt{\\}
    \tcp{Initialization}
    Create $N$ random policies $\{\pi_{\theta_{i}}\}_{i=\{1,N\}}$\;
    Evaluate and insert them in $\mathbb{M}$
    
    \texttt{\\}
    \tcp{Main loop}
    $iteration\_number \xleftarrow{} 0$\;
    \While{$iteration\_number < I$}{
        \texttt{\\}
        \tcp{Sampling and mutation}
        Sample N policies $\{\pi_{\theta_i}\}_{i = 1,N}$ in repertoire $\mathbb{M}$\;
        Mutate half the policies using the TD3 update \cite{fujimoto2018addressing} using $Q_v$\;
        Mutate the other half with random genetic mutations\;

        \texttt{\\}
        \tcp{Train the critic}
        Sample batches of transitions in replay buffer $\mathbb{B}$\;
        Update the critic $Q_v$\ using TD3 \cite{fujimoto2018addressing}
            
        \texttt{\\}
        \tcp{Evaluation}
        Evaluate each new policy over $1$ trajectory and store all transitions in buffer $\mathbb{B}$\;
        Compute their fitnesses and BDs

        \texttt{\\}
        \tcp{Insertion in repertoire}
        For each new policy, insert it in $\mathbb{M}$ only if its fitness is higher than the corresponding policy already in $\mathbb{M}$

        \texttt{\\}
        $iteration\_number = iteration\_number + 1$
    }

    \caption{PGA-MAP-Elites}
    \label{alg:PGAME}
\end{algorithm}

\subsection{PGA-MAP-Elites Low-Spread}
\label{sec:pgame_lowspread_pseudocode}

The PGA-MAP-Elites Low-Spread (PGAME-LS) algorithm is analogous to ME-LS and simply modifies the PGAME algorithm to include an additional constraint over the policies spread (see Equation~\ref{eq:spread_computation}) during the insertion phase. Its overall structure is identical to the standard PGAME algorithm except for the fact that PGAME-LS evaluates solutions over multiples trajectories and insert new solutions into the repertoire only if they present higher fitnesses \textit{and} lower spreads than their corresponding solutions in the repertoire. Algorithm~\ref{alg:PGAME-LS} shows the pseudocode for PGAME-LS.

\makeatletter
\makeatother

\SetKwComment{Comment}{/* }{ */}

\SetArgSty{textnormal}

\begin{algorithm}
    \small
    \SetAlgoLined
    \DontPrintSemicolon
    \SetKwInput{KwInput}{Given}
    \KwInput{
    \begin{itemize}
        \item Max iteration number $I$
        \item Number of evaluations per solution $E$
        \item Sample size $N$
        \item MAP-Elites repertoire $\mathbb{M}$
        \item Replay Buffer $\mathbb{B}$
        \item A critic network $Q_v$
    \end{itemize}
    }
    
    \texttt{\\}
    \tcp{Initialization}
    Create $N$ random policies $\{\pi_{\theta_{i}}\}_{i=\{1,N\}}$\;
    Evaluate and insert them in $\mathbb{M}$
    
    \texttt{\\}
    \tcp{Main loop}
    $iteration\_number \xleftarrow{} 0$\;
    \While{$iteration\_number < I$}{
        \texttt{\\}
        \tcp{Sampling and mutation}
        Sample N policies $\{\pi_{\theta_i}\}_{i = 1,N}$ in repertoire $\mathbb{M}$\;
        Mutate half the policies using the TD3 update \cite{fujimoto2018addressing} using $Q_v$\;
        Mutate the other half with random genetic mutations\;

        \texttt{\\}
        \tcp{Train the critic}
        Sample batches of transitions in replay buffer $\mathbb{B}$\;
        Update the critic $Q_v$\ using TD3 \cite{fujimoto2018addressing}
            
        \texttt{\\}
        \tcp{Evaluation}
        Evaluate each new policy over $E$ trajectories and store all transitions in buffer $\mathbb{B}$\;
        Compute each policy's fitness as its avg. fitness over the $E$ trajectories\;
        Compute each policy's BD as its most frequent BD over the $E$ trajectories
        
        \texttt{\\}
        \tcp{Insertion in repertoire}
        For each new policy, insert it in $\mathbb{M}$ only if its fitness is higher and its spread is lower than the corresponding policy in $\mathbb{M}$

        \texttt{\\}
        $iteration\_number = iteration\_number + 1$
    }

    \caption{PGA-MAP-Elites Low-Spread}
    \label{alg:PGAME-LS}
\end{algorithm}

\subsection{Quality-Diversity Transformer}
\label{sec:qdt_pseudocode}
The Quality-Diversity Transformer model, training method and evaluation method are described in Algorithm~\ref{alg:QDT}. The QDT takes sequences of conditioning BD (which stays the same along the whole trajectory), observations and actions, and produces actions for all time steps in the trajetory in once inference. During training, it allows to compute actions for entire trajectories at once and compare them against labels. Since the QDT is a causal Transformer (GPT-based), for any given time step, it can only attend to all elements that precede this time step, which allows to run inference on whole trajectories without cheating. During evaluation, we simply take the predicted action that corresponds to the current time step and feed it to the environment to obtain the new state. We autoregressively build sequences of conditioning BD, observations and actions that are given as input to the QDT at each time step until the end of the episode. 

\makeatletter
\makeatother

\SetKwComment{Comment}{/* }{ */}

\SetArgSty{textnormal}

\begin{algorithm}
    \small
    \SetAlgoLined
    \DontPrintSemicolon
    \SetKwInput{KwInput}{Given}
    \KwInput{
    \begin{itemize}
        \item Target BD, Observations, Actions:  $BD, O, A$
        \item Causal Transformer model (GPT) $Transformer$
        \item Embedding layers for each modality: $E_{BD}, E_O, E_A$
        \item Time step embedding layer $E_t$
        \item Linear action prediction layer $Pred_A$
        \item Episode length $T$
    \end{itemize}
    }
    
    \texttt{\\}
    \tcp{QDT model}
    \textbf{def} $QDT(BD, O, A, t)$:\;
        \Indp \tcp{Compute inputs embeddings}
        timestep\_emb = $E_t(t)$\;
        BD\_emb = $E_{BD}(BD) + $ timestep\_emb\;
        O\_emb = $E_O(O) + $ timestep\_emb\;
        A\_emb = $E_A(A) + $ timestep\_emb\;

        \tcp{Interleave inputs as $(BD, O_1, A_1 , ... , BD, O_T)$}
        inputs\_emb = interleave(BD\_emb, O\_emb, A\_emb)
   
        \tcp{Use the transformer to process inputs}
        hidden\_states = $Transformer$(inputs\_emb)

        \tcp{Predict actions}
        return $Pred_A$(hidden\_states)

    \texttt{\\}
    \Indm \tcp{Training loop}
    \tcp{Dims: (batch\_size, $T$ , dim)}
    \For{$(BD, O, A , t)$ \textbf{in} dataloader}{
    A\_preds = $QDT(BD, O, A, t)$\;
    loss = mean((A\_preds - A)**2)\;
    optimizer.zero\_grad(); loss.backward(); optimizer.step()
    }

    \texttt{\\}
    \tcp{Evaluation loop (autoregressive generation)}
    $BD$ = generate\_target\_BD( )\;
    $BD, O, A, t,$ done = [$BD$], [env.reset( )], [ ], [$1$] , False\;
    \While{not done}{
        \tcp{Sample next action}
        action = $QDT(BD, O, A, t)$[$t$]\;
        new\_O, done = env.step(action)\;
        \tcp{Append new elements to sequence}
        $BD = BD +$[$BD$]\;
        $O, A, t$ = $O$+[new\_O], $A$+[action] , $t$+[len($BD$)]

    }    

    \caption{Quality-Diversity Transformer}
    \label{alg:QDT}
\end{algorithm}

\section{Hyperparameters}
\label{sec:hyperparameters}
In this section we present hyperparameters used in our experiments. For each algorithm presented, we used the same hyperparameters for both environments.

As mentioned in Appendix~\ref{sec:algo_pseudocodes}, we based this work on the QDax framework \cite{lim2022accelerated, chalumeau2023qdax} and used hyperparameters values presented in \cite{chalumeau2022neuroevolution} for MAP-Elites and PGA-MAP-Elites (as well as their Low-Spread counterparts), which are standard values in the literature.
Considering that QD algorithms and their Low-Spread versions share almost all their hyperparameters, we include them in the same tables, the only additional hyperparameter in ME-LS and PGAME-LS being the number of times each solution is evaluated, which is $10$ for all settings. Table~\ref{table:hp_mapelites} presents hyperparameters for MAP-Elites and MAP-Elites Low-Spread, and Table~\ref{table:hp_pgame} presents hyperparameters for PGA-MAP-Elites and PGA-MAP-Elites Low-Spread. Note that the environment batch size corresponds to the number of solutions that are evolved at each iteration, as we take advantage of the parallelization capabilities of QDax.

\begin{table}[h!]
    \begin{center}
    \begin{tabular}{lc}
        \toprule
        \textbf{Hyperparameter} & \textbf{Value} \\
        \midrule
        Environment batch size & $1000$ \\
        Policy hidden layers size & $[256,256]$\\
        Iso sigma &  $0.005$ \\
        Line sigma &  $0.05$ \\
        \bottomrule
    
    \end{tabular}
    \end{center}
    \caption{Hyperparameters for MAP-Elites and MAP-Elites Low-Spread.}
    \label{table:hp_mapelites}
\end{table}

\begin{table}[h!]
    \begin{center}
    \begin{tabular}{lc} 
        \toprule
        \textbf{Hyperparameter} & \textbf{Value} \\
        \midrule
        Environment batch size & $1000$ \\
        Policy learning rate &  $0.001$\\
        Critic learning rate &  $0.0003$\\
        Policy hidden layers size &  $[256,256]$\\
        Critic hidden layers size &  $[256,256]$\\
        
        Policy noise & $0.2$\\
        Noise clip & $0.5$\\
        Discount & $0.99$\\
        Reward scaling & $1.0$\\
        
        Policy gradient proportion & $50\%$ \\
        Critic training steps & $300$\\
        Policy training steps & $100$\\
        Iso sigma & $0.005$ \\
        Line sigma & $0.05$ \\
        \bottomrule
    
    \end{tabular}
    \end{center}
    
    \caption{Hyperparameters for PGA-MAP-Elites and PGA-MAP-Elites Low-Spread.}
    \label{table:hp_pgame}
\end{table}

\begin{table}[ht!]
    \begin{center}
    \begin{tabular}{lc}
        \toprule
        \textbf{Hyperparameter} & \textbf{Value} \\
        \midrule
        Number of layers & $4$ \\
        Number of attention heads & $8$\\
        Embedding dimension &  $256$ \\
        Nonlinearity function &  ReLU \\
        Batch size & $256$ \\
        Dropout & $0.1$ \\
        Learning rate & $0.0007$ \\

        \bottomrule
    
    \end{tabular}
    \end{center}
    \caption{Hyperparameters for the Quality-Diversity Transformer.}
    \label{table:hp_qdt}
\end{table}

\section{Additional Results}
\label{sec:additional_results}
In this section we present additional results that could not be included in the main paper. We first show that the reproducibility problem holds for PGA-MAP-Elites and that similar observation can be made in Halfcheetah-Uni for both MAP-Elites and PGA-MAP-Elites. Secondly we present traditional QD training metrics for ME, PGAME, ME-LS and PGAME-LS and show that although these training metrics are significantly in favor of the original versions (ME and PGAME), they do not capture the true quality of final repertoires. To support these claim we present a reassessment experiment in which we evaluate all repertoires post-training. We also present the results of the accuracy experiment of the main paper for PGAME variants and of the generalization experiment in Halfcheetah-Uni. Finally, we present evaluation fitness results of the QDT in both environment and demonstrate that it achieves fitnesses that are in line with the true repertoire fitnesses shown in Table~\ref{tab:metrics_xp}.

\subsection{The Reproducibility Problem}
\label{sec:add_results_qd_problem}
\begin{figure*}[ht]
\centering
    \includegraphics[width=1\textwidth]{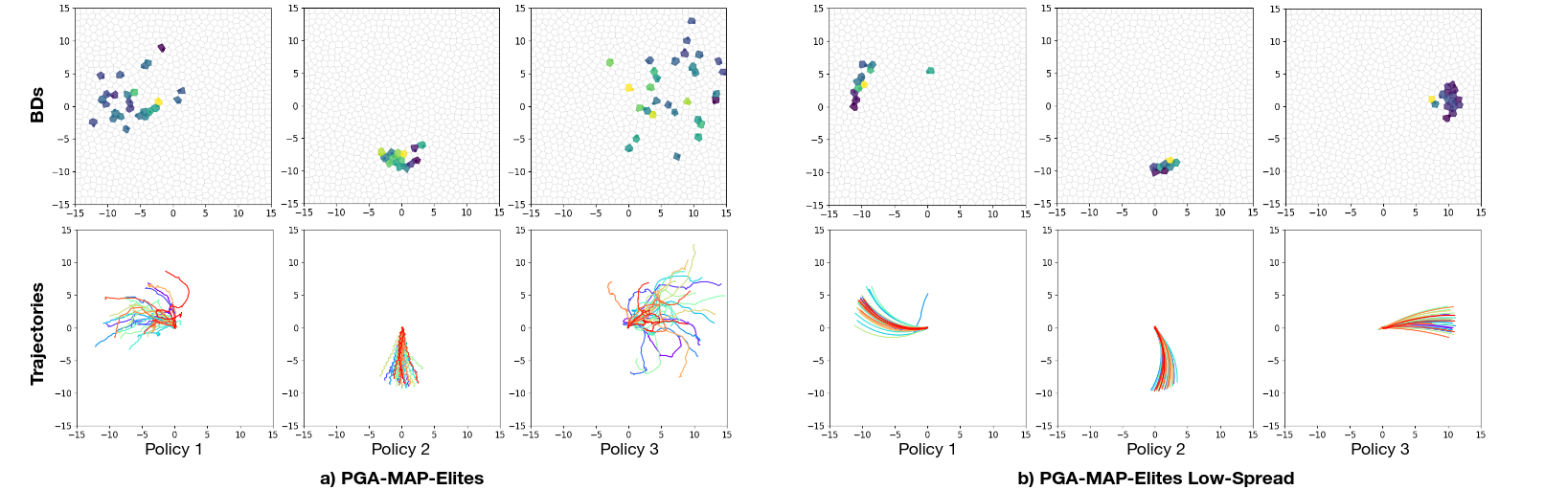}
    \caption{Illustration of the reproducibility problem in Ant-Omni. We select 3 representative policies from final repertoires that have been generated by a) PGA-MAP-Elites and our proposed variant b) PGA-MAP-Elites Low-Spread, and play 30 episodes with each policy using varying initial states. The top row depicts the final BDs obtained by each policy and the bottom row represents the corresponding entire trajectories in the behavior space.}
    \label{fig:add_qd_problem}
\end{figure*}

Figure~\ref{fig:add_qd_problem}a illustrates the reproducibility problem in Ant-Omni for PGAME. We selected $3$ representative policies from a final repertoire produced by PGAME and ran multiple ($N=30$) episodes with each policy. Results are similar to ME policies in that they generate irregular trajectories and demonstrate high spread in the behavior space even though PGAME incorporates a policy-gradient-based mutation operator during its training process. Figure~\ref{fig:add_qd_problem}b show that the Low-Spread version of PGAME, PGAME-LS, does not suffer from these problems and produces policies that are consistent in the BD space and which produce smooth, regular trajectories.

\begin{figure*}[ht]
\centering
    \includegraphics[width=1\textwidth]{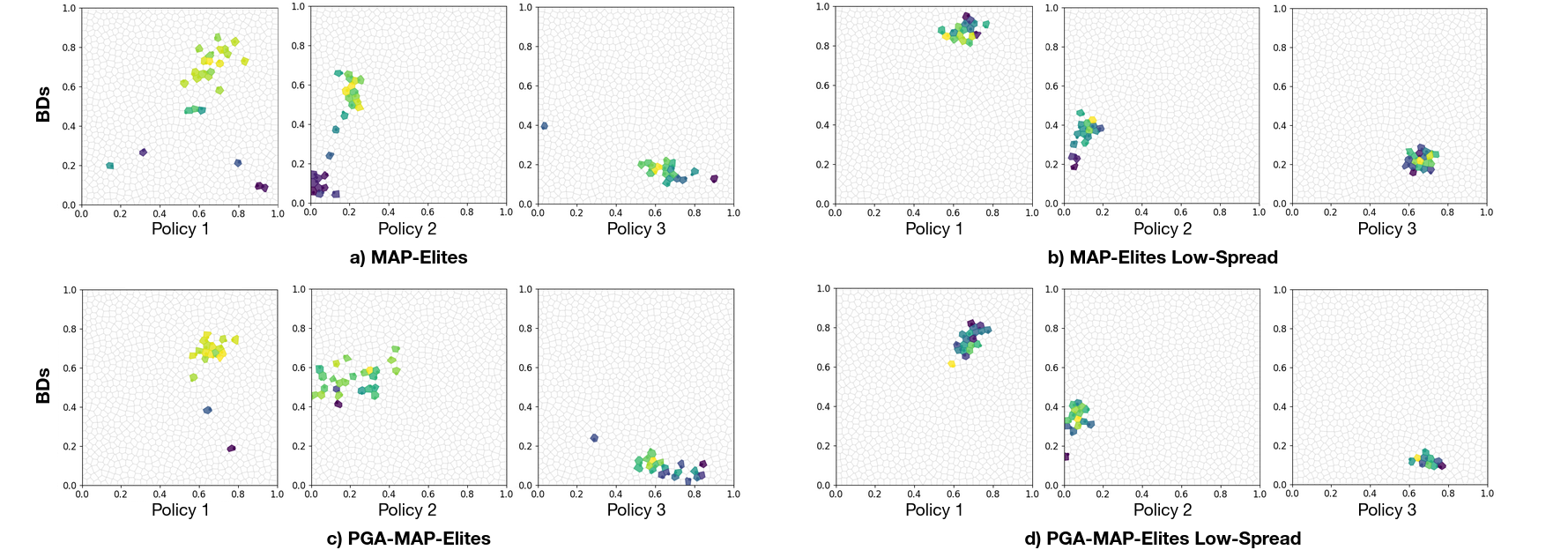}
    \caption{Illustration of the reproducibility problem in Halfcheetah-Uni. We select 3 representative policies from final repertoires that have been generated by a) MAP-Elites, b) MAP-Elites Low-Spread, c) PGA-MAP-Elites, d) PGA-MAP-Elites Low-Spread and play 30 episodes with each policy using varying initial states. Plots depict the final BDs obtained by each policy. Contrary to Ant-Omni in Figures~\ref{fig:qd_problem} and \ref{fig:add_qd_problem} we do not show entire trajectories since it would not be relevant considering the different nature of the behavior space in Halcheetah-Uni.}
    \label{fig:add_qd_problem_halfcheetah}
\end{figure*}

We reproduce these experiments and show in Figure~\ref{fig:add_qd_problem_halfcheetah} that the same observation can be made in the Halfcheetah-Uni environment for both algorithm families. Original versions of these algorithms (ME and PGAME) produce solutions that display high spread in the BD space while their Low-Spread counterparts create consistent solutions. For this environment we do not show whole trajectories as the behavior space is of 
a different nature which is not suitable for such plots.

\subsection{QD Algorithms Results}
\label{sec:add_results_qd_algos}

\begin{figure*}[ht]
\centering
    \includegraphics[width=1\textwidth]{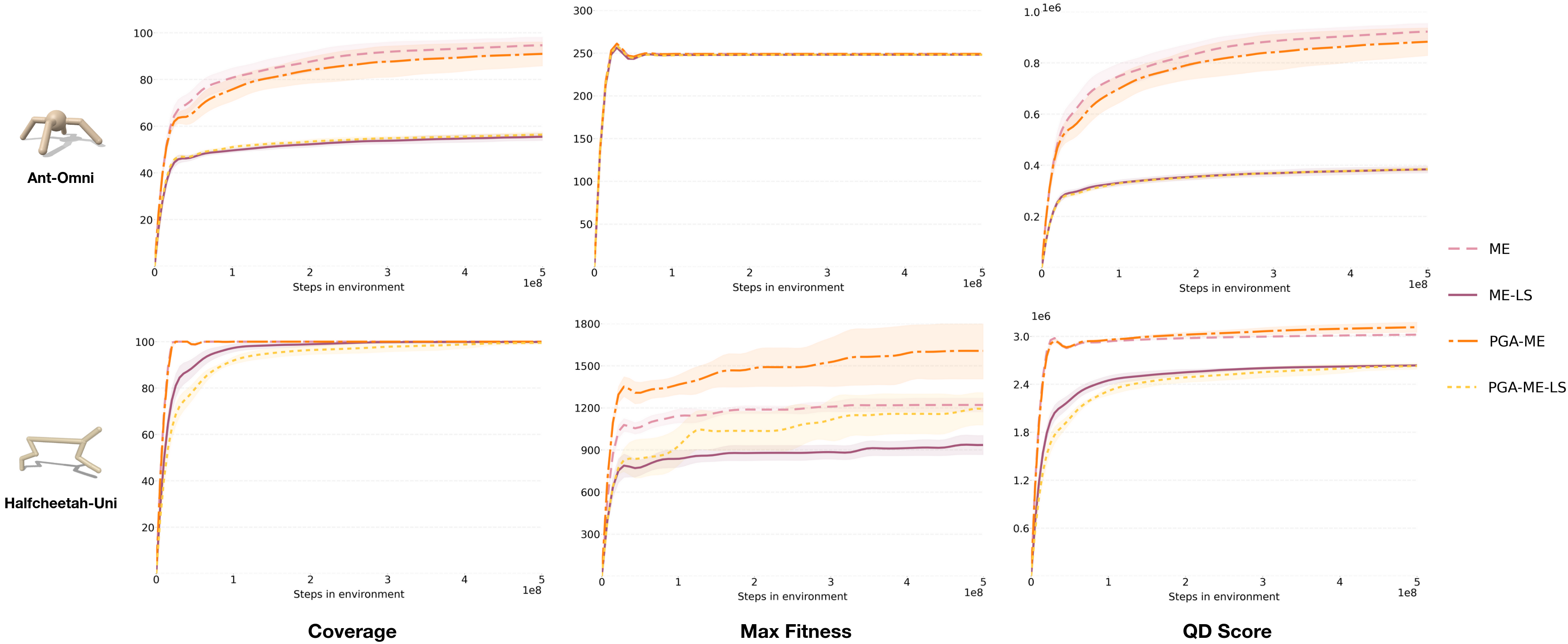}
    \caption{Results of the Quality-Diversity algorithms: MAP-Elites (ME), PGA-MAP-Elites (PGAME) and their Low-Spread variations (ME-LS and PGAME-LS) in both environments over 5 seeds. Coverage indicates the proportion of the behavior space that have been covered in the repertoire, max fitness reports the best fitness obtained by any solution evaluated so far and the QD score represents the total sum of fitness across all solutions in the repertoire. Performances are plotted against the number of interactions with the environment.}
    \label{fig:add_results_qd_algos}
\end{figure*}

\begin{figure*}[ht]
\centering
    \includegraphics[width=1.0\textwidth]{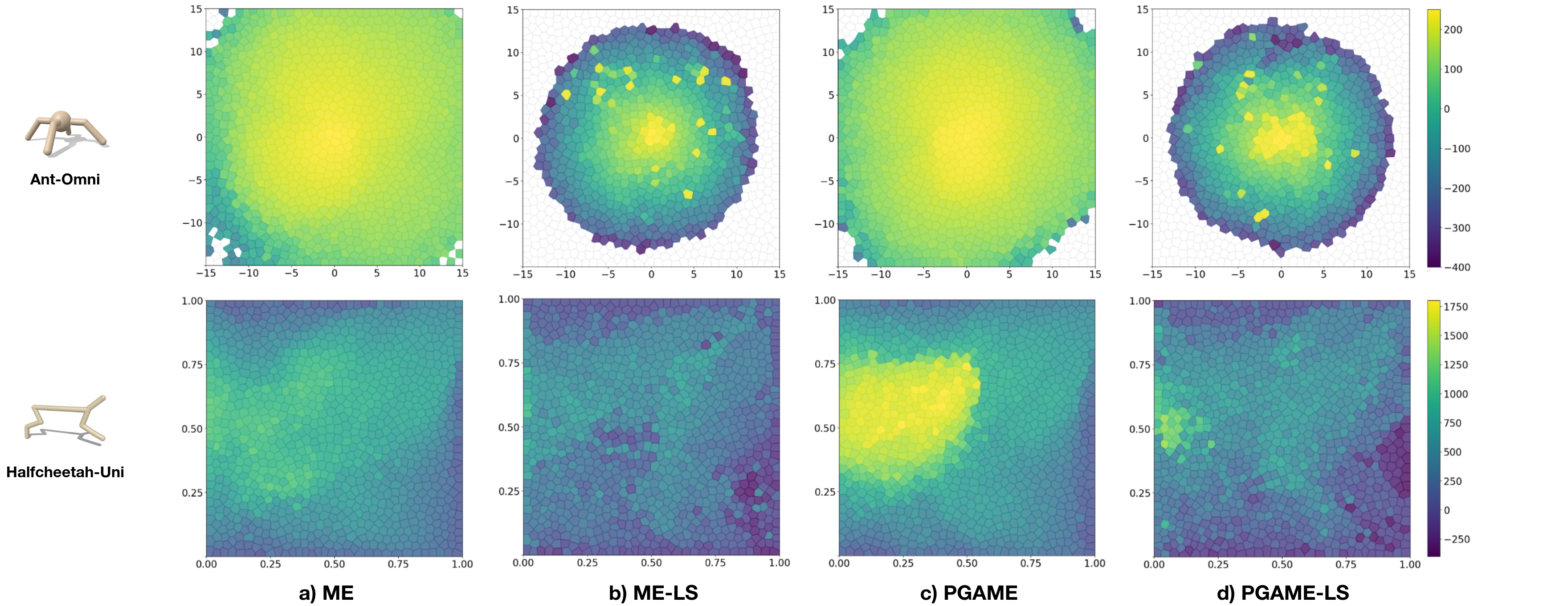}
    \caption{Coverage maps of the Quality-Diversity algorithms: a) MAP-Elites, b) MAP-Elites Low-Spread, c) PGA-MAP-Elites and d) PGA-MAP-Elites Low-Spread. Fitness is represented by color: lighter is better.}    \label{fig:add_results_coverage_maps}
\end{figure*}

Figure~\ref{fig:add_results_qd_algos} and Figure~\ref{fig:add_results_coverage_maps} gather results for the QD algorithms runs, namely MAP-Elites (ME), PGA-MAP-Elites (PGAME) and their Low-Spread versions (ME-LS and PGAME-LS), and show their respective coverages,  maximum fitnesses, and QD scores in both environments.
In accordance with standard practices in QD research, we add an offset to the fitnesses when computing the QD score to ensure that it is an increasing function of the coverage. The initial repertoires, which are identical for all methods, are of size 1024 and are generated using Centroidal Voronoi Tessellations \cite{DBLP:journals/corr/VassiliadesCM16}.

Results show that the original versions of these algorithms (ME and PGAME) obtain the best performances over all training metrics by a significant margin. However, it is important to recall that first, the original versions evaluate each policy only once, contrary to the Low-Spread versions that do multiple evaluations (usually $10$), which strongly promotes accidental policies that have been lucky over their unique evaluation episode obtaining abnormally high fitnesses and inaccurate BDs. And second, The Low-Spread versions include an additional insertion criteria that drives the search process towards policies that are more consistent in the BD space, leading to fewer insertions in the repertoire --policies have to show higher average fitness and better consistency than their counterpart in the repertoire to be selected.
We argue that these very points are responsible for the difference in coverage and, \textit{in fine}, in maximum fitness and QD score. But more importantly, even though the original versions present better training metrics, they remain limited as these metrics do not take into account the actual usefulness of a QD repertoire and its capacity to deliver accurate and consistent solutions that behave according to their BDs, as shown in Section~\ref{sec:reproductibility_problem} and Figure~\ref{fig:results_precision}.

\begin{table}[!ht]
    \small
    \caption{Results of the reassessment experiment in Halfcheetah-Uni. For each algorithm, we take a final repertoire of policies and test them again over multiple episodes. We insert them into a new, empty repertoire and report its coverage, max fitness and QD score. "Initial" columns show values for the initial repertoire, that is, the repertoire that was used during the algorithm run. "Recalc." columns refer to values of the new repertoire that contains solutions after re-evaluation. It appears that after re-evaluation the repertoires issued from Low-Spread methods show superior performance compared to the original methods.
    }
        \centering
        \label{tab:metrics_xp}
        \begin{tabular}{ccccccc}
            \toprule
            
             & \multicolumn{2}{c}{Coverage}  & \multicolumn{2}{c}{Max Fitness} & \multicolumn{2}{c}{QD Score}\\
             & \multicolumn{2}{c}{(in \%)}  & \multicolumn{2}{c}{} & \multicolumn{2}{c}{$(\times10^6)$}\\
                & Initial & {\bf Recalc.} & Initial & {\bf Recalc.} & Initial &{\bf Recalc.}\\
            \midrule
            ME & $100$ &  ${\bf 43}$ & $1226$ & ${\bf 770}$ & $3.05$ & ${\bf 1.17}$\\
            ME-LS & $100$ & ${\bf 55}$ & $992$ & ${\bf 730}$ & $2.63$ & ${\bf 1.41}$\\
            PGAME & $100$ & ${\bf 40}$ & $1417$ & ${\bf 977}$ & $3.07$ & ${\bf 1.10}$\\
            PGAME-LS & $100$ & ${\bf 53}$ & $1194$ & ${\bf 1073}$ & $2.63$ & ${\bf 1.36}$\\
            \bottomrule
        \end{tabular}
\end{table}

\begin{figure*}[h]
\centering
    \includegraphics[width=0.95\textwidth]{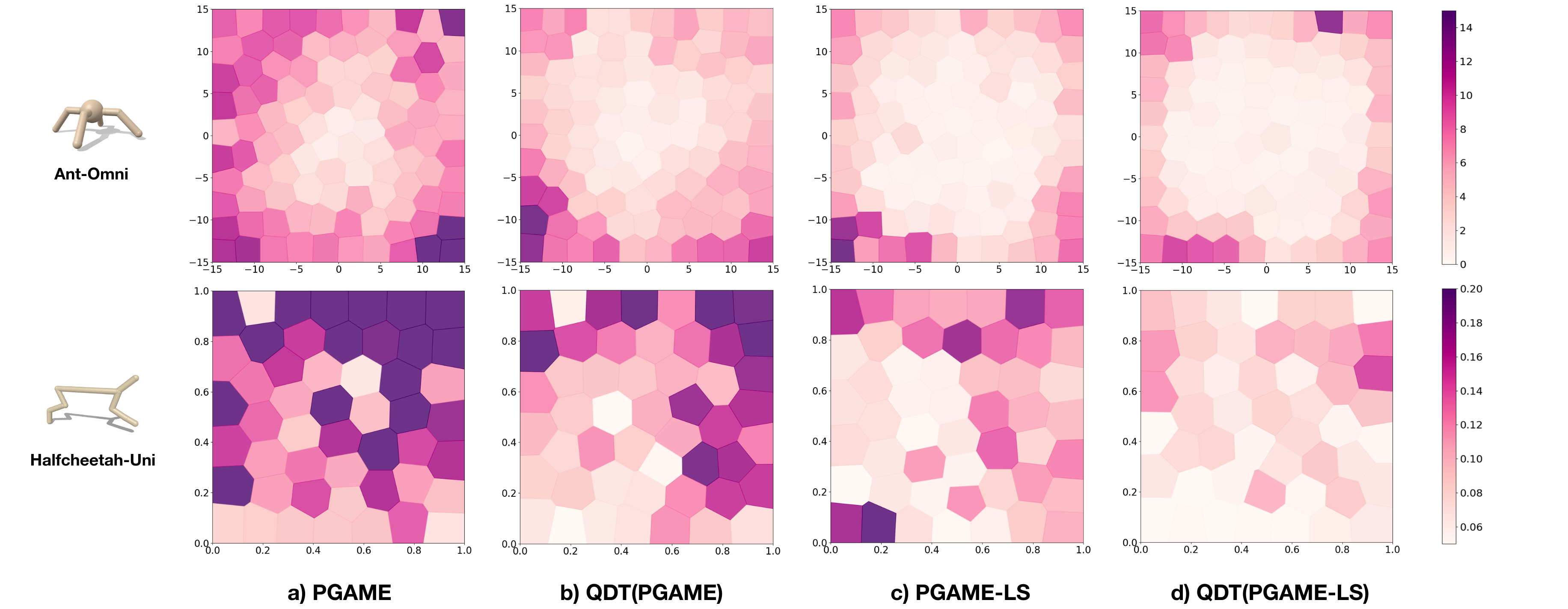}
    \caption{Results of the accuracy experiment for the PGAME variants. This experiment can be described in 2 steps: 1. We select multiple evaluation goals (target BDs) in the behavior space, 100 and 50 for Ant-Omni and Halfcheetah-Uni respectively. To get meaningful goals, we simply compute a CVT of the BD space in which goals are the centers of each zone, 2. For each goal, we play 10 episodes and plot their average Euclidean distance to the goal. For PGAME and PGAME-LS, trajectories are played by the nearest policy to the goal in the repertoire. For the QDT, we simply condition it on the goal. Distance is represented by color: lighter is better. The QDT(PGAME-LS) appears to be the most accurate method to achieve behaviors on demand.}
    \label{fig:results_precision_pgame}
\end{figure*}

\subsubsection{The Reassessment Experiment}
\label{sec:reassessment_xp}
To further prove this statement, we take a final repertoire of each method --namely ME, ME-LS, PGAME and PGAME-LS-- and conducted the following experiment: 1. We take each policy from the repertoire and play 10 episodes with varying initial states, 2. We compute the policy's average fitness and its most frequent behavior, 3. We add evaluated policies into a new, empty repertoire according to their recalculated fitnesses and behaviors. Table~\ref{tab:metrics_xp} gathers results for this experiment for the Halfcheetah-Uni environment. It is clear that after re-evaluation, ME-LS and PGAME-LS demonstrate superior coverages and QD scores to ME and PGAME, and comparable max fitnesses. This experiment shows that for uncertain domains, training metrics such as those usually presented in the QD literature (see Figure~\ref{fig:add_results_qd_algos}) can be misleading and may not capture true value of a final repertoire. Note that similar results concerning MAP-Elites have been observed in different settings \cite{engebraaten2020framework}.

\subsection{Accuracy Experiment}
\label{sec:add_accuracy_xp}

Figure~\ref{fig:results_precision_pgame} depicts the accuracy experiment described in Section~\ref{sec:accuracy_xp} for PGAME variants. We evaluate each method against multiple evaluation goals (target BDs) that reasonably cover the behavior space and report the average distance for each goal, which is represented by color (lighter is better). Similar to results presented in Section~\ref{sec:accuracy_xp}, PGAME fails to achieve target BDs on demand, while the QDT(PGAME) improves over this result. The QDT(PGAME-LS) appears to be the most accurate method to achieve target BDs in both environments. Importantly, note that all methods struggle to reach the most outer goals in Ant-Omni, this is due to the fact that no policy --hence no data-- is available for these zones of the BD space as shown in the dataset representation in Figure~\ref{fig:results_generalization}.

\subsection{Generalization Experiment}
\label{sec:add_generalization_experiment}

\begin{figure*}[h]
\centering
    \includegraphics[width=1\textwidth]{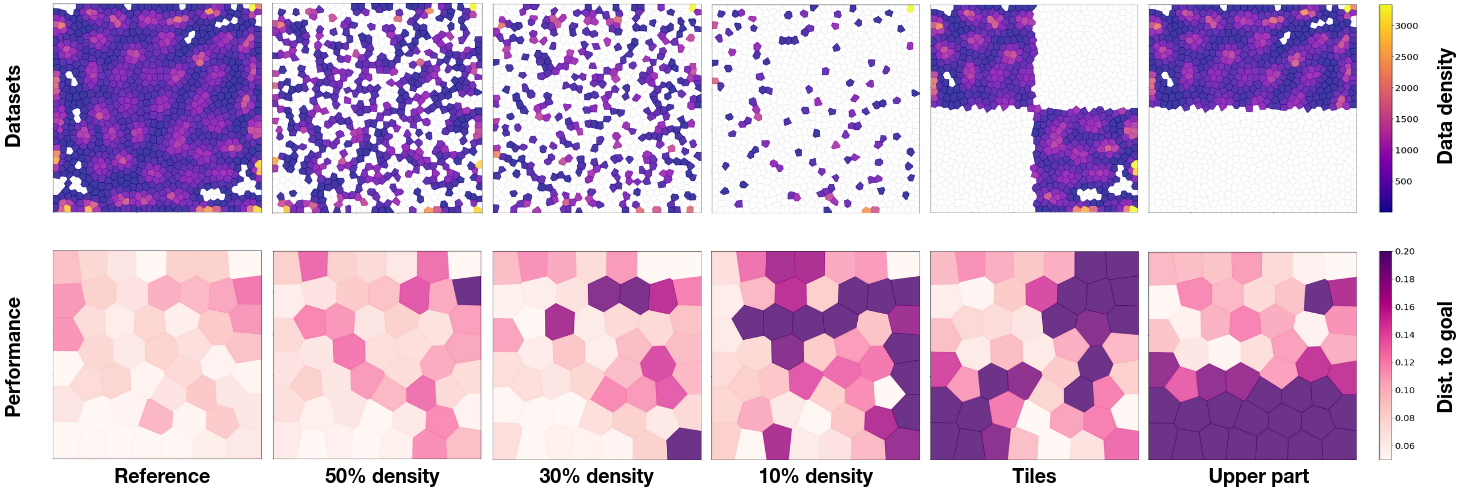}
    \caption{Results of the QDT generalization experiment in Halfcheetah-Uni. In this experiment we run accuracy experiments (bottom row) on truncated datasets (top row) which are deprived of a part of their trajectories. The QDT shows strong interpolation ability on the 50\%, and 30\% density datasets and a limited ability to extrapolate in "Tiles" and "Upper part" datasets where entire zones of the BD space are deprived of data.}
    \label{fig:add_results_generalization}
\end{figure*}

Figure~\ref{fig:add_results_generalization} depicts the generalization experiment described in Section~\ref{sec:generalization_experiment} for the Halfcheetah-Uni environment. We observe that, even though the QDT demonstrates good accuracy up to the 30\% density setting, it has more difficulties to generalize in this environment, both for interpolation and extrapolation. We hypothesize that this difference between Ant-Omni and Halfcheetah-Uni comes from the very different nature of their behavior spaces. After execution of a QD algorithm in Ant-Omni, two policies that are close in the BD space often produce similar full-body trajectories, meaning that they walk on the 2D plane and reach their final positions, which happens to be slightly different, but both policies walk with similar gaits. In Halfcheetah-Uni, two policies that are close in the BD space can demonstrate radically different full-body behaviors. As an extreme example, it occurred that we observed neighboring policies in the BD space, one of which was doing backflips while the other was running normally. We believe that these gaps in real behaviors prevent effective generalization for the QDT. Finally, note that in these generalization experiments we simply prune trajectories from the datasets and do not increase the number of trajectories in preserved zones, which can affect the model in the sense that it has strictly less data to train on.

\subsection{QDT Fitness Results}
\label{sec:qdt_fitness}

\begin{figure*}[ht]
    \includegraphics[width=0.6\textwidth]{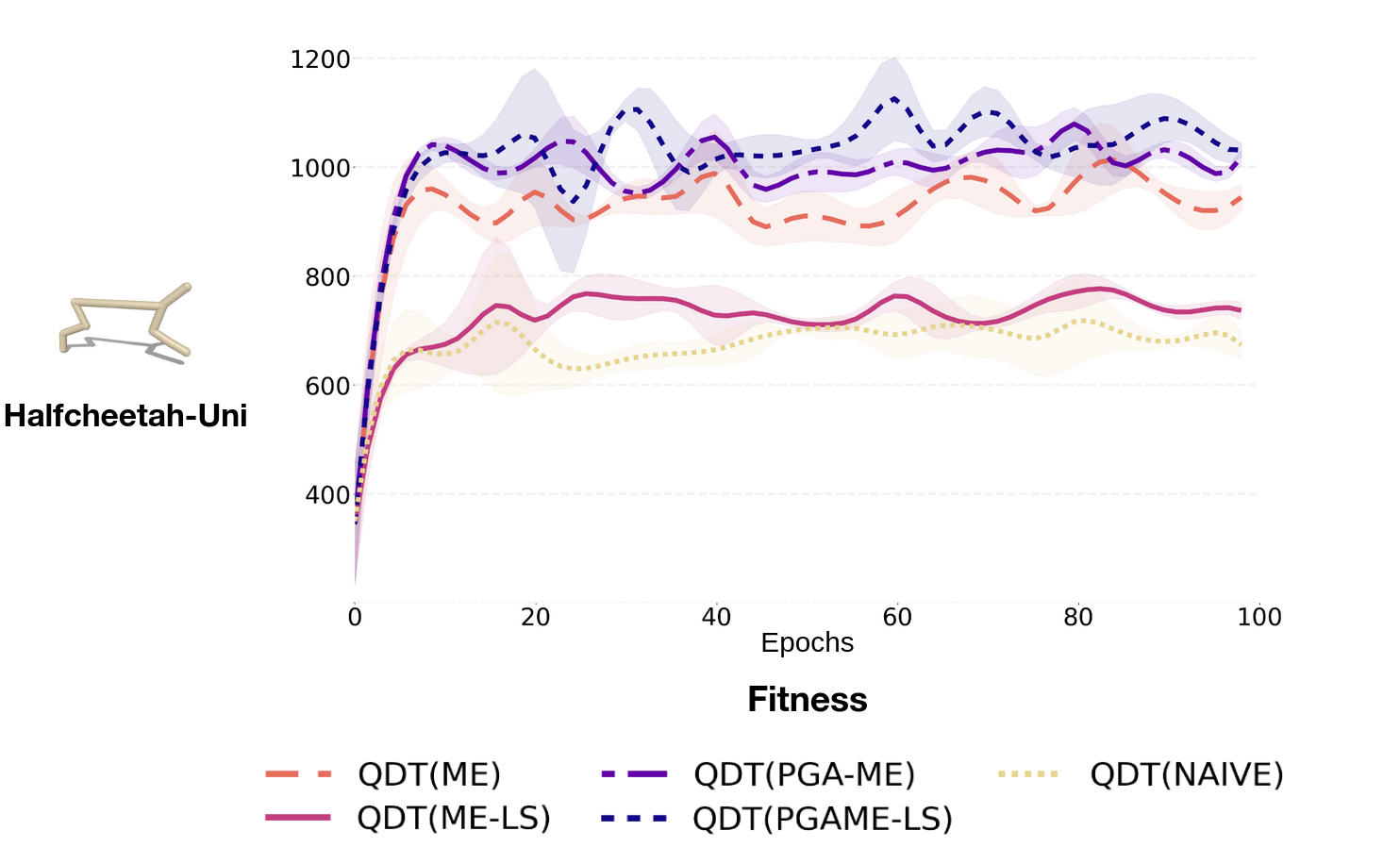}
    \caption{Maximum fitness of the QDT in Halfcheetah-Uni for evaluations during the training phase (average values and std ranges on 3 seeds). We report the maximum fitness obtained over all goals. Performances are similar --if not superior-- to values reported in the reassessment experiment in Table~\ref{tab:metrics_xp}, meaning that our model is able to replicate the fitness of QD policies.}
    \label{fig:qdt_training_fitness}
\end{figure*}

Figure~\ref{fig:qdt_training_fitness} reports the performances of all variants of the QDT in terms of fitness for Halfcheetah-Uni. During evaluation phases of the training process, which are described in Section~\ref{sec:training_and_ablations}, we record the average fitness obtained for each goal (target BD). The maximum fitness reported in Figure~\ref{fig:qdt_training_fitness} simply corresponds to maximum over all goals. To be fair, these results should be compared to results of the reassessment experiment in Table~\ref{tab:metrics_xp} as we want to know what is the maximum average fitness that we can expect from each method at evaluation time. It appears that the QDT is able to reproduce the maximum fitnesses of the QD policies that were used to generate its dataset.

\end{document}